%% file: main_arxiv.tex
\pdfoutput=1
\documentclass[10pt,twocolumn,letterpaper]{article}

\usepackage[pagenumbers]{cvpr}
\usepackage[T1]{fontenc}

\usepackage{amsmath,amssymb}
\usepackage{booktabs,tabularx,multirow}
\usepackage{pifont}
\newcommand{\cmark}{\ding{51}}
\newcommand{\xmark}{\ding{55}}
\newcommand{\pmark}{$\circ$}
\usepackage{algorithm,algpseudocode}
\usepackage{tikz,xcolor,graphicx,float}
\usepackage{capt-of}
\usepackage{pgfplots}
\usepgfplotslibrary{groupplots}
\usetikzlibrary{calc}
\pgfplotsset{compat=1.18}
\definecolor{cBlue}{HTML}{5F91EB}
\definecolor{cMint}{HTML}{78D4B7}
\definecolor{cPurple}{HTML}{9F7EDF}
\definecolor{cTeal}{HTML}{61B6CA}
\definecolor{cOrange}{HTML}{EEA64F}
\definecolor{cPink}{HTML}{E47CAC}
\definecolor{cInk}{HTML}{333333}
\definecolor{cMuted}{HTML}{5F5E58}
\definecolor{cNeutral}{HTML}{7D7C75}
\definecolor{cGray}{HTML}{C9C8C0}
\definecolor{cGrid}{HTML}{E4E3DC}
\definecolor{cSand}{HTML}{D2B48C}
\definecolor{cSage}{HTML}{94BF8A}
\definecolor{cRose}{HTML}{DC8F8C}
\colorlet{cKiwi}{cSand}
\colorlet{cCoinve}{cSage}
\colorlet{cHthree}{cRose}
\colorlet{cAgent}{cBlue}
\pgfplotsset{
  mtpanel/.style={width=0.348\textwidth, height=4.1cm, grid=major, grid style={cGrid},
    tick label style={font=\scriptsize}, label style={font=\small}, title style={font=\small, yshift=-3pt},
    axis line style={cMuted}, tick style={cMuted},
    ylabel shift=-5pt, yticklabel style={inner sep=1pt}, xlabel shift=-3pt,
    every axis plot/.append style={line width=1pt, mark size=1.5pt}},
  mtkiwi/.style={cKiwi, mark=square*}, mtcoinve/.style={cCoinve, mark=triangle*},
  mththree/.style={cHthree, mark=diamond*}, mtagent/.style={cAgent, mark=*, line width=1.6pt},
}
\usepackage{listings}
\usepackage{microtype}
\usepackage{placeins}
\usepackage{enumitem}
\newlist{tightitemize}{itemize}{1}
\setlist[tightitemize]{label=\textbullet, leftmargin=1.2em, itemsep=1pt, topsep=2pt, parsep=0pt}
\usetikzlibrary{arrows.meta,positioning,fit,calc}

\definecolor{cvprblue}{rgb}{0.21,0.49,0.74}
\usepackage[pagebackref,breaklinks,colorlinks,allcolors=cvprblue]{hyperref}

\lstdefinestyle{prompt}{
  basicstyle=\ttfamily\scriptsize,
  breaklines=true,
  frame=single,
  columns=fullflexible,
  keepspaces=true,
  showstringspaces=false
}
\newcommand{\benchname}{ChatVE-Bench}
\newcommand{\agentname}{VideoBanana}
\newcommand{\supvideos}{}
\input{generated/numbers}


\definecolor{rb1}{HTML}{E53935}\definecolor{rb2}{HTML}{F4511E}\definecolor{rb3}{HTML}{FB8C00}
\definecolor{rb4}{HTML}{F2A900}\definecolor{rb5}{HTML}{7CB342}\definecolor{rb6}{HTML}{43A047}
\definecolor{rb7}{HTML}{00897B}\definecolor{rb8}{HTML}{039BE5}\definecolor{rb9}{HTML}{1E88E5}
\definecolor{rb10}{HTML}{5E35B1}\definecolor{rb11}{HTML}{8E24AA}
\definecolor{weblabel}{HTML}{D81B60}
\newcommand{\rainbowname}{\texorpdfstring{\textcolor{rb1}{V}\textcolor{rb2}{i}\textcolor{rb3}{d}%
\textcolor{rb4}{e}\textcolor{rb5}{o}\textcolor{rb6}{B}\textcolor{rb7}{a}\textcolor{rb8}{n}%
\textcolor{rb9}{a}\textcolor{rb10}{n}\textcolor{rb11}{a}}{VideoBanana}}

\title{\rainbowname: Agentic Multi-Turn Conversational Video Editing}
\author{
Jiayi Zhang$^{1}$\quad
Xiangbo Gao$^{2}$\quad
Vardhan Dongre$^{3}$\\
Franck Dernoncourt$^{4}$\quad
Ryan A. Rossi$^{5}$\quad
Zhengzhong Tu$^{2\star}$\\[1mm]
{\normalsize $^{1}$The University of Texas at Austin\quad
$^{2}$Texas A\&M University\quad
$^{3}$University of Illinois Urbana-Champaign}\\
{\normalsize $^{4}$Adobe Research\quad
$^{5}$Zillow}\\[1mm]
{\small $^{\star}$Corresponding Author: \texttt{tzz@tamu.edu}.\quad
Contact: \texttt{jz28724@eid.utexas.edu}.}\\
{\small \textcolor{weblabel}{\textbf{Project Website}}:
\href{https://jiayizhanga.github.io/VideoBanana/}{\texttt{jiayizhanga.github.io/VideoBanana}}}
}

\begin{document}
\twocolumn[{%
\renewcommand\twocolumn[1][]{#1}%
\maketitle
\begin{center}
\includegraphics[width=\textwidth]{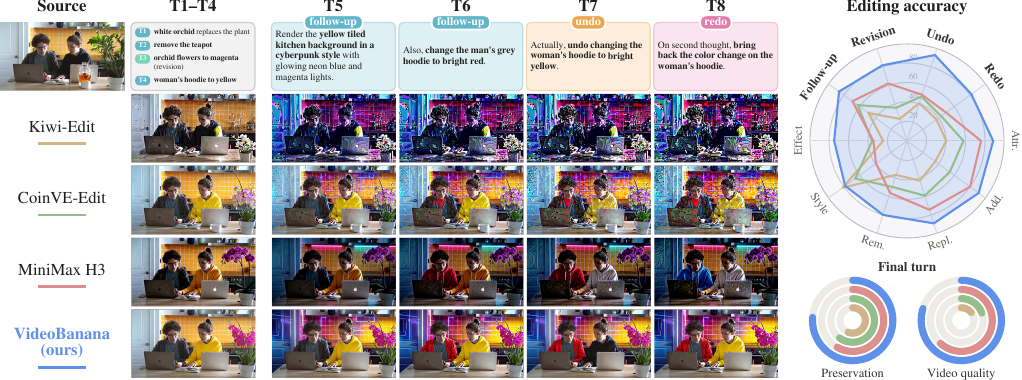}
\captionof{figure}{Multi-turn video editing on \benchname{}. \agentname{} follows the conversation
better than turn-by-turn editing baselines. Left: one session. The baselines edit their own outputs.
Kiwi-Edit and CoinVE-Edit lose video quality, and no baseline restores the black hoodie when T7 undoes
T4. \agentname{} restores it, keeps the later edits, and brings the yellow back at T8. Right: editing
accuracy by action and edit type, and final-turn preservation and video quality, over all sessions.
\agentname{} is ahead on almost every axis and in the final turn.}
\label{fig:teaser}
\end{center}
\vspace{2mm}
}]

\begin{abstract}
Video editing is often multi-turn and conversational. Within one session, a user gives instruction
after instruction, and each result should reflect all of them so far. Each instruction depends on the
conversation history, so it can revise or undo an earlier edit.
Instruction-guided video editors, however, are built for a single request on the source video, and
existing video editing benchmarks score each request on its own. We therefore introduce \textbf{\benchname{}},
a benchmark of multi-turn, conversational video editing. Its \nSessions{} sessions of 5 to 8 turns add
follow-up edits, revise earlier edits, undo any earlier step, and redo undone edits. Every turn is
scored with a checklist built from the expected state of the session. On \benchname{}, editors used turn-by-turn
edit their own outputs again and again. After a few generations, their video quality falls, and so
does the accuracy of new edits. They can undo only by issuing another edit, which often fails to
restore what the undone edit replaced. We then propose
\textbf{\agentname{}}, an agent that treats such editing as tracking the goal of the conversation over a tree
of verifier-scored checkpoints. At every turn, it restates the full goal as atomic intents and chooses the
entry video from any earlier checkpoint, so undo becomes backtracking and every output stays within a
few generations of the source. It writes editor instructions for the intents that entry does not yet
satisfy, executes them with a single-turn editing agent, and verifies the output against the goal.
With the same editor,
\agentname{} raises final-turn editing accuracy from \nFinalIFCoinve{} to \nFinalIFAgent{} and content
preservation from \nFinalCPCoinve{} to \nFinalCPAgent{} over editing turn by turn. It undoes
reliably and largely keeps video quality close to the source at every turn.
We will release \benchname{}, its evaluation code, and \agentname{} at \href{https://github.com/taco-group/VideoBanana}{\texttt{github.com/taco-group/VideoBanana}}.
\end{abstract}

\begin{figure*}[t]
\centering
\includegraphics[width=\textwidth]{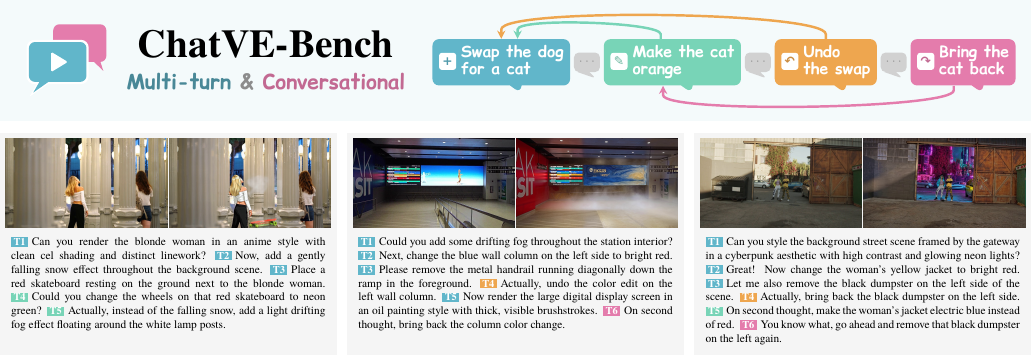}
\caption{Examples from \benchname{}. Each panel shows the source video, \agentname{}'s final output,
and every request of the session.}
\label{fig:bench-examples}
\end{figure*}
\input{sections/introduction}
\begin{table*}[t]
\centering
\small
\caption{Comparison with existing video editing benchmarks. \emph{Conversational}: some requests are
conversation-dependent. \emph{State-based}: each turn is scored against the expected state of the
whole session. Checklist questions are counted at every turn they appear in.}
\label{tab:bench-compare}
\setlength{\tabcolsep}{0pt}
\newcommand{\twoline}[2]{\begin{tabular}[b]{@{}c@{}}#1\\#2\end{tabular}}
\begin{tabular*}{\textwidth}{@{\extracolsep{\fill}}lcccccccccc@{}}
\toprule
& & \multicolumn{2}{c}{Turns} & \multicolumn{2}{c}{Checklist questions} & & & & & \\
\cmidrule(lr){3-4}\cmidrule(lr){5-6}
Benchmark & Sessions & \twoline{per}{session} & total & \twoline{per}{session} & total & \twoline{Multi-}{turn} & \twoline{Conver-}{sational} & Undo & Revision & \twoline{State-}{based} \\
\midrule
VEFX-Bench~\cite{vefxbench} & 300 & 1 & 300 & -- & -- & \xmark & \xmark & \xmark & \xmark & \xmark \\
CoinVE-Bench~\cite{coinve200k} & 361 & 1 & 361 & \nStQuestionsPerCase{} & \nStQuestions{} & \xmark & \xmark & \xmark & \xmark & \xmark \\
CoVEBench~\cite{covebench} & 626 & 1 & 626 & 16.0 & 9{,}990 & \xmark & \xmark & \xmark & \xmark & \xmark \\
IVEBench~\cite{ivebench} & 600 & 1 & 600 & -- & -- & \xmark & \xmark & \xmark & \xmark & \xmark \\
\midrule
\textbf{\benchname{} (ours)} & 144 & 5--8 & 907 & \nQuestionsPerSession{} & \nQuestions{} & \cmark & \cmark & \cmark & \cmark & \cmark \\
\bottomrule
\end{tabular*}
\end{table*}
\input{sections/related_work}
\input{sections/benchmark}
\input{sections/method}
\input{sections/experiments}

\input{sections/conclusion}

{
    \small
    \bibliographystyle{ieeenat_fullname}
    \bibliography{agentic_references}
}

\clearpage
\appendix
\input{sections/appendix}
\clearpage
\raggedbottom
\input{prompts_appendix}

\end{document}

%% file: generated/numbers.tex
\newcommand{\nAblCapSessions}{98}
\newcommand{\nAblConcatCP}{74.1}
\newcommand{\nAblConcatCapCP}{73.9}
\newcommand{\nAblConcatCapIF}{77.6}
\newcommand{\nAblConcatIF}{78.4}
\newcommand{\nBhCalls}{1166}
\newcommand{\nBhComposite}{164}
\newcommand{\nBhDepthOne}{433}
\newcommand{\nBhDepthThree}{95}
\newcommand{\nBhDepthTwo}{378}
\newcommand{\nBhDepthZero}{1}
\newcommand{\nBhExhaustedSessions}{13}
\newcommand{\nBhExhaustedTurns}{35}
\newcommand{\nBhGatePass}{658}
\newcommand{\nBhMaxDepth}{3}
\newcommand{\nBhNoReuseUndoEarlier}{55}
\newcommand{\nBhRedo}{77}
\newcommand{\nBhReuseInPlace}{3}
\newcommand{\nBhReuseRedo}{30}
\newcommand{\nBhReuseUndoLast}{52}
\newcommand{\nBhSecond}{344}
\newcommand{\nBhSecondPass}{180}
\newcommand{\nBhUndoEarlier}{58}
\newcommand{\nBhUndoLast}{56}
\newcommand{\nCallRatio}{1.29}
\newcommand{\nCtlEndedPctOff}{56}
\newcommand{\nCtlEndedPctOn}{97}
\newcommand{\nCtlEndedQ}{95}
\newcommand{\nCtlGoalActiveSessions}{72}
\newcommand{\nCtlVerifActiveTurns}{252}
\newcommand{\nDepthCountCPN}{142}
\newcommand{\nDepthCountCPOneHigher}{55}
\newcommand{\nDepthCountCPThreeHigher}{50}
\newcommand{\nDepthCountEditsThreeHigher}{295}
\newcommand{\nDepthCountFrameThreeHigher}{313}
\newcommand{\nDepthCountIFOneHigher}{38}
\newcommand{\nDepthCountIFSame}{66}
\newcommand{\nDepthCountIFThreeHigher}{40}
\newcommand{\nDepthCountSPAN}{131}
\newcommand{\nDepthEntryEarlierEditTurns}{197}
\newcommand{\nDepthEntryEarlierTurns}{210}
\newcommand{\nDepthEntryPrevEditTurns}{189}
\newcommand{\nDepthEntryPrevTurns}{219}
\newcommand{\nDepthLaterTurns}{763}
\newcommand{\nDepthOneIF}{84.0}
\newcommand{\nDepthRegionTurns}{386}
\newcommand{\nDepthSessions}{144}
\newcommand{\nDepthStoredTurns}{429}
\newcommand{\nDepthThreeIF}{85.3}
\newcommand{\nDevClipSessions}{8}
\newcommand{\nDevClips}{10}
\newcommand{\nDistinctQuestions}{2{,}805}
\newcommand{\nDistinctQuestionsPerSession}{19.5}
\newcommand{\nFinalAbsPCAgent}{0.385}
\newcommand{\nFinalCPAgent}{76.2}
\newcommand{\nFinalCPCoinve}{60.3}
\newcommand{\nFinalCPHthree}{58.4}
\newcommand{\nFinalDVQRabsAgent}{1}
\newcommand{\nFinalDVQRabsCoinve}{76}
\newcommand{\nFinalIFAgent}{85.3}
\newcommand{\nFinalIFCoinve}{50.6}
\newcommand{\nFinalIFHthree}{63.8}
\newcommand{\nFinalPCFromSource}{0.383}
\newcommand{\nFinalQuestions}{2{,}301}
\newcommand{\nFinalQuestionsPerSession}{16.0}
\newcommand{\nFinalSPAAgent}{87.7}
\newcommand{\nFinalSPACoinve}{42.2}
\newcommand{\nGateEditFailPct}{33}
\newcommand{\nGateEditPct}{78}
\newcommand{\nGateEditsPerTurn}{1.5}
\newcommand{\nGateJudged}{819}
\newcommand{\nGateKeptPerTurn}{1.1}
\newcommand{\nGateOnlyEditPct}{55}
\newcommand{\nGateOnlyRestPct}{7}
\newcommand{\nGateRejected}{341}
\newcommand{\nGateRestFailPct}{11}
\newcommand{\nGateRestOneBelow}{170}
\newcommand{\nGateRestPct}{26}
\newcommand{\nHeldFinalIFAgent}{85.1}
\newcommand{\nHeldFinalIFCoinve}{50.6}
\newcommand{\nHeldSessions}{136}
\newcommand{\nOEdFinalCPKiwiAgent}{65.7}
\newcommand{\nOEdFinalCPKiwiTbt}{55.9}
\newcommand{\nOEdFinalIFKiwiAgent}{71.8}
\newcommand{\nOEdFinalIFKiwiTbt}{35.3}
\newcommand{\nOEdSecondCoinve}{430}
\newcommand{\nOEdSecondKiwi}{584}
\newcommand{\nOneStepBackCPDev}{76.9}
\newcommand{\nOneStepBackCPFirst}{82.3}
\newcommand{\nOneStepBackIFDev}{81.0}
\newcommand{\nOneStepBackIFFirst}{79.8}
\newcommand{\nOneStepBackSADev}{71.3}
\newcommand{\nOneStepBackSAFirst}{79.8}
\newcommand{\nOneStepBackTurns}{59}
\newcommand{\nOneStepCPDev}{78.7}
\newcommand{\nOneStepCPFirst}{84.4}
\newcommand{\nOneStepIFDev}{81.8}
\newcommand{\nOneStepIFFirst}{85.5}
\newcommand{\nOneStepReuseCPDev}{80.1}
\newcommand{\nOneStepReuseCPFirst}{86.1}
\newcommand{\nOneStepReuseIFDev}{82.5}
\newcommand{\nOneStepReuseIFFirst}{89.5}
\newcommand{\nOneStepReuseSADev}{76.6}
\newcommand{\nOneStepReuseSAFirst}{88.4}
\newcommand{\nOneStepReuseTurns}{82}
\newcommand{\nOneStepSADev}{74.3}
\newcommand{\nOneStepSAFirst}{84.7}
\newcommand{\nOneStepTurns}{141}
\newcommand{\nOpRevAgent}{81.6}
\newcommand{\nOpRevBaseHi}{62.0}
\newcommand{\nOpRevBaseLo}{23.8}
\newcommand{\nOpTurnsUndoEarlier}{58}
\newcommand{\nOpTurnsUndoLast}{56}
\newcommand{\nOpUndoAllAgent}{92.7}
\newcommand{\nOpUndoAllBaseHi}{49.8}
\newcommand{\nOpUndoAllBaseLo}{40.8}
\newcommand{\nOpUndoEarlierAgent}{98.4}
\newcommand{\nOpUndoLastCoinve}{66.0}
\newcommand{\nOpWholeUndoEarlierIFAgent}{84.4}
\newcommand{\nOpWholeUndoLastIFCoinve}{66.8}
\newcommand{\nQuestions}{10{,}903}
\newcommand{\nQuestionsPerSession}{75.7}
\newcommand{\nQwenAllIFAgent}{85.3}
\newcommand{\nQwenAllIFGapAbs}{1.1}
\newcommand{\nQwenAllIFQwen}{84.2}
\newcommand{\nQwenDevSessions}{40}
\newcommand{\nQwenHeldSessions}{104}
\newcommand{\nQwenNote}{}
\newcommand{\nRedoFailedTurnsAgent}{4}
\newcommand{\nRedoIFHeldAgent}{80.8}
\newcommand{\nRedoIFHeldBaseMax}{26.5}
\newcommand{\nRedoIFHeldBaseMin}{10.9}
\newcommand{\nRedoIFHeldLenientBaseMax}{26.6}
\newcommand{\nRedoIFHeldLenientBaseMin}{14.5}
\newcommand{\nRedoUndoQMax}{2}
\newcommand{\nRedoUndoQMin}{1}
\newcommand{\nRevertOKAgent}{232}
\newcommand{\nRevertOKCoinve}{76}
\newcommand{\nRevertQ}{246}
\newcommand{\nSessions}{144}
\newcommand{\nSessionsTurnEight}{27}
\newcommand{\nSessionsTurnSeven}{62}
\newcommand{\nSessionsTurnSix}{98}
\newcommand{\nSourcePC}{0.461}
\newcommand{\nStAvgAgent}{91.1}
\newcommand{\nStAvgCoinve}{89.1}
\newcommand{\nStCases}{361}
\newcommand{\nStIdentical}{293}
\newcommand{\nStPCAgent}{0.41}
\newcommand{\nStQuestions}{4{,}131}
\newcommand{\nStQuestionsPerCase}{11.4}
\newcommand{\nStRejected}{103}
\newcommand{\nStRepAvgCoinve}{90.9}
\newcommand{\nStRerunGap}{1.7}
\newcommand{\nStRpCases}{45}
\newcommand{\nStVQMaxDiff}{0.04}
\newcommand{\nSurvCoinveFinal}{82.8}
\newcommand{\nSurvEdits}{103}
\newcommand{\nSurvKiwiFinal}{71.9}
\newcommand{\nSurvKiwiOne}{82.5}
\newcommand{\nTQBelowFromCoinve}{5}
\newcommand{\nTQBelowFromKiwi}{4}
\newcommand{\nTurnDVQRAgentMaxAbs}{2}
\newcommand{\nTurnDVQRabsCoinveFive}{57}
\newcommand{\nTurnIFAgentMax}{91.5}
\newcommand{\nTurnIFAgentMin}{83.3}
\newcommand{\nTurnIFCoinveFive}{56.6}
\newcommand{\nTurnOneDTQCoinve}{+14}
\newcommand{\nTurnOneDTQHthree}{+6}
\newcommand{\nTurnOneGapCoinve}{2.8}
\newcommand{\nTurnOnePCAgent}{0.447}
\newcommand{\nTurns}{907}
\newcommand{\nTurnsWithActionPct}{37.8}
\newcommand{\nTypeMadeAddAgent}{93.6}
\newcommand{\nTypeMadeAddCoinve}{62.3}
\newcommand{\nTypeMadeEffectAgent}{73.6}
\newcommand{\nTypeMadeEffectCoinve}{43.8}
\newcommand{\nTypeMadeStyleAgent}{78.4}
\newcommand{\nTypeMadeStyleCoinve}{67.1}
\newcommand{\nTypeMadeStyleKiwi}{91.3}
\newcommand{\nTypeTurnEditsAdd}{23/44/28}
\newcommand{\nTypeTurnEditsAttr}{28/40/34}
\newcommand{\nTypeTurnEditsEffect}{28/34/26}
\newcommand{\nTypeTurnEditsRemove}{22/39/31}
\newcommand{\nTypeTurnEditsRepl}{19/49/30}
\newcommand{\nTypeTurnEditsStyle}{24/36/29}
\newcommand{\nVerifAllFinalIFOff}{80.7}
\newcommand{\nVlmCalls}{4{,}077}
\newcommand{\nVlmCompile}{1{,}185}
\newcommand{\nVlmPerTurn}{4.5}
\newcommand{\nVlmRefine}{907}
\newcommand{\nVlmVerify}{1{,}162}
\newcommand{\nVlmWrite}{823}

%% file: sections/introduction.tex
\section{Introduction}
\label{sec:intro}

Instruction-guided video editors now follow detailed requests such as replacing an object,
recoloring a garment, adding an effect, or restyling a whole clip~\cite{ditto,kiwiedit,sama}. Their training
data and benchmarks share one assumption: a single request is applied to the source video in
one pass (e.g.,~\cite{senorita,ditto,coinve200k,vefxbench,covebench}). Real editing is interactive. Within one session, a user gives instruction after
instruction and inspects each result. A later instruction may add another element, revise an earlier
choice (``actually, make it dark blue instead''), undo it (``undo the change that made it blue''), or
restore something from the original video. Such editing is \emph{multi-turn}, because
each result should reflect all instructions so far. It is also \emph{conversational}, because each
instruction depends on the conversation history. It often cannot be carried out from its own words and
the current video alone. The same editors are therefore used turn by turn, and three problems appear that
single-request evaluation does not expose.

\noindent\textbf{Generative editing is lossy.}
Even a single generative image edit often makes changes that were not requested~\cite{collins2026everyday}.
Chaining an editor on its own outputs re-encodes the whole clip at every turn. On our benchmark,
video quality holds for a few generations and then falls quickly, along with editing accuracy
(Sec.~\ref{sec:exp-quality}).

\noindent\textbf{Generative editing is hard to reverse.}
Undoing an edit by issuing another edit asks the editor to recreate content it has already
overwritten. Chained editors fail to restore most of what an undo asks for
(Secs.~\ref{sec:exp-main} and~\ref{sec:exp-ops}).

\noindent\textbf{Earlier failures carry over unnoticed.}
An edit that a turn did not achieve stays missing, and later turns that build on it fail as well: a
revision cannot change an element that was never made (Sec.~\ref{sec:exp-ops}). A system that remembers
only its requests, not what the videos show, builds on these failures.

We address these problems with \agentname{}, an agent that treats multi-turn, conversational editing as tracking the goal of
the conversation over a tree of verifier-scored checkpoints (Fig.~\ref{fig:mt-architecture}). After
each turn \agentname{} restates the full goal as atomic intents; a goal that returns to an
earlier state reuses the earlier intent. Each output becomes a checkpoint that records the
intents the verifier found in it. For each turn, \agentname{} picks the entry video from the whole
tree: undo becomes backtracking to the latest checkpoint without the unwanted edit and a depth limit
keeps outputs close to the source. It then writes instructions for the intents the entry does not yet
show, executes them with a single-turn editing agent, and verifies the result against the goal.

\noindent\textbf{Contributions.}
\begin{itemize}
    \item A formulation of multi-turn, conversational video editing whose target is the state set by
          all turns so far, with explicit revision, undo, and redo (Secs.~\ref{sec:bench-taxonomy}
          and~\ref{sec:problem}).
    \item \textbf{\benchname{}}, a benchmark of 5- to 8-turn sessions that revise, undo, and redo earlier edits,
          with each turn scored against the expected state of the session (Sec.~\ref{sec:benchmark}).
    \item \textbf{\agentname{}}, an agent that verifies every output against the goal of the
          conversation, undoes by backtracking over a tree of verifier-scored checkpoints, and keeps
          outputs shallow, with an unchanged single-request editor (Sec.~\ref{sec:method}).
    \item Experiments with the same editor: final-turn editing accuracy rises from \nFinalIFCoinve{} to
          \nFinalIFAgent{} over turn-by-turn editing, with better preservation and video quality
          (Sec.~\ref{sec:experiments}).
\end{itemize}

%% file: sections/related_work.tex
\section{Related Work}
\label{sec:related}

\noindent\textbf{Instruction-guided video editing.}
Current video editors train on large instruction datasets~\cite{insv2v,senorita,ditto,coinve200k}
(other designs in App.~\ref{app:bench-compare-full}). Their benchmarks score each request on its
own~\cite{vefxbench,ivebench,openve,five,coinve200k,covebench}. We keep CoinVE-Bench's question formats and evaluator but
compose every turn's checklist from the state of the whole conversation. Multi-turn video editing is little studied. Existing methods retrieve earlier outputs as context~\cite{memoryv2v} or edit
streams of generated segments or shots~\cite{infinityedit,guo2026contextmaster}, and commercial models
edit video through conversation~\cite{deepmind2026geminiomniflash}. These change or condition
the editor. Multi-turn image editors map the dialogue to executable editing commands~\cite{kim2022caise}, treat undo as a dialogue intent~\cite{lin2018dialogedit}, train on
rollbacks~\cite{zhang2026nonmarkov}, or re-edit the initial image at every
turn~\cite{cui2023chatedit}. \agentname{} keeps the editor unchanged and returns to or
continues from verified outputs.

\noindent\textbf{Multi-turn editing benchmarks.}
Multi-turn editing has been benchmarked mostly on
images~\cite{magicbrush,edival,lmebench,imgedit,editr2,agentbanana}. For video,
Sora100K~\cite{yang2026sora100k} scores chains of edits mined from a creation platform. Each edit is
defined relative to its input video. To our knowledge, \benchname{} is the first
multi-turn benchmark for editing a given video whose turns depend on the conversation and are scored against the session state.

\noindent\textbf{Agents for visual editing.}
LLM and multimodal agents decompose a request, call generators or editing tools, and check the
result~\cite{huang2026agenticvisualgen}, also for multi-turn image
editing~\cite{shen2026imagagent,ma2026talk2image} (more in App.~\ref{app:bench-compare-full}). Verifiers that check an output and
retry~\cite{madaan2023selfrefine,shinn2023reflexion,yang2024idea2img,lee2026videorepair} judge it
against a single request. In a conversation, an output must also keep the
edits of earlier turns, and later turns need a record of what each output shows. Our verifier therefore
checks the goal of the whole conversation and keeps its scores as the state of each checkpoint.
Existing agents keep earlier outputs as context or for traceability~\cite{liang2025univa}. Our checkpoint
tree indexes them by the intents verified in them, so an undo returns to an existing node. Unlike search or
planning trees~\cite{yao2023tot,zhou2024lats,wang2024genartist}, it stores committed outputs and does not expand candidate steps.

%% file: sections/benchmark.tex
\section{\benchname{}: Benchmarking Multi-Turn Conversational Video Editing}
\label{sec:benchmark}

We present \benchname{}, a benchmark for multi-turn conversational video editing. It contains
\nSessions{} sessions and \nTurns{} turns on clips from VEFX-Bench~\cite{vefxbench}; each session
starts from one video and spans 5--8 turns (6.3 on average). At each turn the user makes one
request, which in \nTurnsWithActionPct{}\% of the turns revises, undoes, or redoes an earlier edit,
and the system returns an edited video, scored against the expected state of the whole session.
\benchname{} measures one ability, following the conversation. Everything else is kept at a level that
current editors already handle: short clips, appearance edits, and one request per turn. It therefore
admits most single-request editors (App.~\ref{app:bench-scope}).
Figure~\ref{fig:bench-examples} shows three sessions (nine in App.~\ref{app:benchmark}),
Figure~\ref{fig:bench-overview} their
composition (statistics in Appendix~\ref{app:bench-stats}), and Table~\ref{tab:bench-compare} the
comparison with existing video editing benchmarks, among which \benchname{} is the only multi-turn one
(image benchmarks in App.~\ref{app:bench-compare-full}).

\begin{figure}[t]
\centering
\includegraphics[width=\columnwidth]{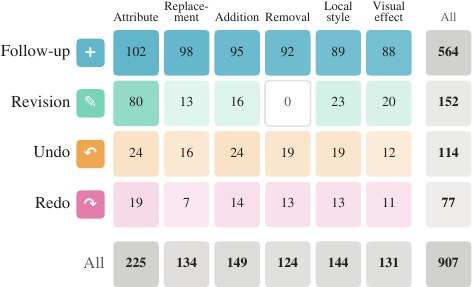}
\caption{\benchname{} by action (rows) and edit type (columns). Each cell is the number of turns
with that action and edit type.}
\label{fig:bench-overview}
\end{figure}

\subsection{Construction}
\label{sec:bench-construction}

We construct sessions with explicit target states at every turn. For each session, a program
creates a structured \emph{blueprint}: the selected edits, the action at each turn, and the expected
state of every edit after that turn. The expected states are the ground truth and depend on neither
the generated wording nor any system output. The pipeline has four steps (App.~\ref{app:bench-steps}).
We collect candidate edits on distinct targets in CoinVE-Bench's atomic-edit
format~\cite{coinve200k}. We schedule revisions, undos, and redos and compute the expected
states. We write a question set for each possible state of an edit
(Sec.~\ref{sec:bench-eval}). Finally, an LLM writes one user message per turn from the
blueprint, simulating the user~\cite{ni2026usersim}. We review the edits, questions, and messages against the source clip with AI assistance,
and an author resolves every judgment call (App.~\ref{sec:bench-review}).

\subsection{Task Taxonomy}
\label{sec:bench-taxonomy}

We label every turn with an edit type and a multi-turn action (\emph{action} for short), and use both
labels to analyze performance separately by action and by edit type (Sec.~\ref{sec:exp-ops}).

\noindent\textbf{Edit types.}
The edit type is the visual change a turn is about: an \emph{attribute} change of an existing element
(color, material, pattern), an \emph{addition}, a \emph{replacement} of an element or the background,
a \emph{removal}, a \emph{local style} on one region~\cite{vefxbench}, or a \emph{visual effect} over
the background (snow, fog, rain).

\noindent\textbf{Actions.}
An edit $e=(o,\pi,v)$ sets property $\pi$ of element $o$ to value $v$. Let $A_t$ be the edits that are
\emph{active} after turn $t$. The expected video after turn $t$ is the source with every active edit
applied. It never depends on an earlier output, so an output that failed to show an edit does not
change what later turns expect. Each turn has one of four actions (formal rules in
App.~\ref{app:bench-ops}). A \emph{follow-up} makes a new change of an element that no earlier edit
touched (``turn the car red''). A \emph{revision} changes what an earlier edit made, either its value
(``actually, deep blue instead'') or the element it created (``make the cat you added black''). An
\emph{undo} removes any active edit, not only the latest, together with the edits that build on it
(``take out the bench again, but keep the rest''). A \emph{redo} makes an undone edit active again
(``put the bench back''). An undo or a redo takes the type of the edit it removes or brings back.

\subsection{Evaluation}
\label{sec:bench-eval}

\noindent\textbf{Per-turn checklist.}
The checklist of turn $t$ asks about every edit of the session in the state the blueprint expects
after that turn, together with the session's preservation questions. An edit that is still requested
($e\in A_t$) is checked for what it requests. An edit that is no longer requested is checked for being
gone: its target is back in its source state, or the replaced value no longer appears. Each edit's
questions are written once and asked again at every later turn in which the edit keeps its state. A score change across turns therefore comes from
the videos, not from the questions (App.~\ref{app:bench-states}).

\noindent\textbf{Evaluator and metrics.}
Each turn is scored as a standard CoinVE-Bench case with its unchanged evaluator. The evaluator
receives the questions but not the conversation, so it judges only what the video shows. We report two
of CoinVE-Bench's scores. Editing accuracy (EA) is the mean of semantic accuracy (SA), scope accuracy (SPA),
and edit persistence (EP). Content preservation (CP) is the measure of its Semantic Preservation.
Objective video-quality metrics replace its physical-naturalness questions (Sec.~\ref{sec:exp-setup}).
The final-turn checklist covers every edit of the session in its final expected state, so its
scores are the main results.

%% file: sections/method.tex
\section{Method}
\label{sec:method}

\begin{figure*}[t]
\centering
\includegraphics[width=\textwidth]{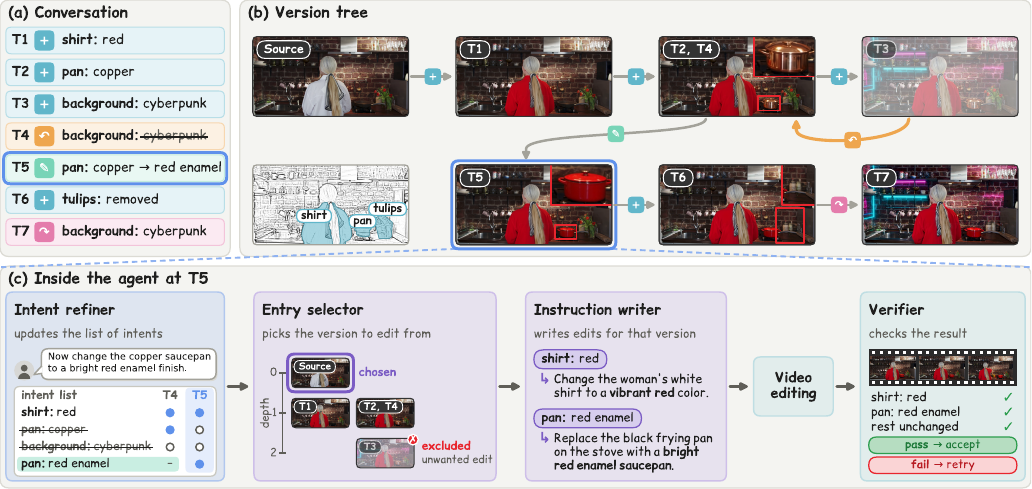}
\caption{Overview of \agentname{}. In (b), T2 and T4 return the same video: when T4 undoes the edit of
T3, a checkpoint that shows the goal already exists and is reused. In (c), at T5, the entry selector does
not edit the output of T3, because it still shows the cyberpunk background that T4 undid. It starts from a
checkpoint without this edit. Among these, it prefers the shallowest output.}
\label{fig:mt-architecture}
\end{figure*}

\agentname{} has three parts around an unchanged editor (Fig.~\ref{fig:mt-architecture} and
Alg.~\ref{alg:conversation}): goal tracking restates what the user wants after every turn
(Sec.~\ref{sec:goal}), a verifier-scored checkpoint tree chooses the video to edit from (Sec.~\ref{sec:tree}),
and a verifier checks each output against the whole goal before it is committed and records what it shows (Sec.~\ref{sec:verify}).

\subsection{Problem Formulation}
\label{sec:problem}

A user edits a source video $V_0$ over turns $u_1,\ldots,u_T$, and after each turn $t$ the system
returns a video $Y_t$. The target of turn $t$ is the \emph{goal} $G_t$: the set of atomic intents that
all turns up to $t$ ask for, each an absolute target state of one element (``the woman's jacket is
electric blue''). $G_t$ follows the active edits of Sec.~\ref{sec:bench-taxonomy}. $Y_t$ should show
every intent in $G_t$, keep all other content of $V_0$, and have high visual quality. The system may
call a single-request editor on any earlier video. We write $d(\cdot)$ for the number of editor calls
between a video and the source $V_0$, its \emph{generation depth}.

\input{figures/algorithm_conversation}

\subsection{Goal Tracking}
\label{sec:goal}

One LLM call per turn restates the goal from the dialogue. The refiner receives every user turn and
every intent recorded so far. It returns the intents that the turn adds, stops, or brings back,
and the complete active list. It never sees the videos, so a failed output cannot
change the goal (App.~\ref{app:goal-entry}).

\noindent\textbf{Reuse.}
When the goal returns to a state that an earlier intent describes, the refiner reuses that intent
instead of writing a new one. Undoing a revision therefore reactivates exactly the intent it replaced,
and the checkpoints that already show it become valid entries again. This lets \agentname{} undo by
going back rather than by editing.

\subsection{Checkpoint Tree and Entry Selection}
\label{sec:tree}

The source and every committed output are nodes of a tree. Each node $n$ stores its video, depth $d(n)$, and a ledger of verifier scores (Sec.~\ref{sec:verify}). An intent is
\emph{realized} at $n$ if its score reaches the pass level $\tau$. The source realizes nothing.

\noindent\textbf{Roles.}
For a candidate entry $e$, the active intents realized at $e$ are preserved, $P(e)$, and the others must be executed, $A(e)$. An inactive intent that is still realized at $e$ is
\emph{unwanted}, unless an executed intent replaces it: executing ``the jacket is electric blue'' from a
red jacket is an ordinary edit.

\noindent\textbf{Selection.}
An entry is \emph{feasible} if no intent is unwanted at it, so undo is never done by editing.
The source is always feasible. With at most $K$ instructions per call, entry $e$ yields an
output of depth $d(e)+\lceil |A(e)|/K\rceil$, and this must stay within $D$. Among feasible
entries within this limit, one with nothing to execute is returned without editing. Otherwise,
redoing a few earlier intents can give a shallower output. Among entries that need at most
$K_{\mathrm{merge}}$ instructions, \agentname{} takes the shallowest output. If there is
none, it takes the entry with the fewest instructions (App.~\ref{app:entry}).

\noindent\textbf{Execution.}
The editor knows nothing about the conversation. One LLM call writes an instruction for each intent to
execute, phrased from what the entry shows (``change the red jacket to electric blue'' from
turn 2's output, ``change the yellow jacket to electric blue'' from the source). A single-turn editing
agent (App.~\ref{app:single-turn-agent}) sends them to the editor, at most $K$ per call.

\begin{table*}[t]
\centering
\small
\caption{Results on \benchname{}. SA, SPA, and EP are semantic accuracy, scope accuracy, and edit
persistence. EA is editing accuracy. CP is content
preservation. $\Delta$ is the change from the session's source clip. Calls counts editor calls; in the
final-turn block, only those at the final turn. VLM is the model family of the agent's refiner, writer,
and verifier. The video editor is unchanged. Bold: best in each block.\nQwenNote{}}
\label{tab:mt-main}
\setlength{\tabcolsep}{2.5pt}
\begin{tabular*}{\textwidth}{@{\extracolsep{\fill}}lccc@{\hspace{7pt}}cccccccccc@{}}
\toprule
& \multicolumn{4}{c}{Editing accuracy} & Preservation & \multicolumn{7}{c}{Video quality} & \\
\cmidrule(lr){2-5}\cmidrule(lr){6-6}\cmidrule(lr){7-13}
Method & SA & SPA & EP & EA & CP & TQ & $\Delta$TQ & VQR & $\Delta$VQR & PC & AES & MSM & Calls \\
\midrule
\input{generated/tab_mt_main.tex} 
\bottomrule
\end{tabular*}
\end{table*}

\subsection{Verification}
\label{sec:verify}

A verifier for a single request asks whether one instruction was carried out. In a conversation this is
not enough. An edit can succeed and still damage an earlier one, and the next turn has to know what the
video now shows. \agentname{}'s verifier therefore differs in three ways.

First, it checks the whole goal. One VLM call sees the entry and the candidate. It scores every active
intent, not only those of the current request, and also the preservation of the rest of the video and
the visual quality (App.~\ref{app:prompts}).

Second, it treats new and earlier edits differently. The intents executed in this turn, preservation,
and visual quality are judged by their own scores. An intent that the entry already showed is compared
with its score at the entry. It only has to stay undamaged, because this turn did not make it
(App.~\ref{app:gate}). A candidate that is not accepted is resampled, and the turn commits the best
candidate.

Third, its scores are kept as the ledger of the new checkpoint. It starts from the entry's ledger: a
judged intent takes its new score, the others keep theirs, and an intent counts as absent once its
replacement is realized.
Entry selection reads these ledgers to know which checkpoint shows which intents
(Sec.~\ref{sec:tree}). Later turns thus build on what the videos show, not on what was requested.

The verifier's scores thus have two uses. As accept decisions, they reject a candidate and resample it.
As ledgers, they make the checkpoints usable as entries. The checkpoint tree therefore cannot be kept
without the verifier.

%% file: figures/algorithm_conversation.tex
\begin{algorithm}[t]
\caption{One conversation turn. $\tilde X_e$: the entry's frames as the editor receives them;
$s_e$: the entry's ledger; $\kappa$: composite score; $\oplus$: overwrite the entry's ledger with the new
scores; \textsc{Gate}: the accept decision of Sec.~\ref{sec:verify}. The rule for intents that keep failing
is in App.~\ref{app:gate}.}
\label{alg:conversation}
\small
\newcommand{\AlgPart}[1]{\Statex \textcolor{cvprblue}{\textbf{#1}}}
\begin{algorithmic}[1]
\Require checkpoint tree $\mathcal{T}$ (videos, depths, ledgers), recorded intents, user turns $u_{1:t}$
\Ensure output $Y_t$ and the updated tree
\AlgPart{Goal and entry}
\State $G_t \gets \Call{RefineGoal}{u_{1:t}, \text{recorded intents}}$
\State $e, A, P \gets \Call{SelectEntry}{\mathcal{T}, G_t}$
\State \textbf{if} $A=\emptyset$ \textbf{then return} $Y_t \gets X_e$
\AlgPart{Edit and verify}
\State $\Pi \gets \Call{WriteInstructions}{X_e, A, P}$
\For{$k = 1, 2$}
    \State $c_k \gets \Call{SingleTurnAgent}{X_e, \Pi, \text{seed}=k-1}$
    \State $z_k \gets \Call{Verify}{\tilde X_e, c_k, A\cup P}$
    \State \textbf{if} $\Call{Gate}{z_k, s_e}$ \textbf{then break}
\EndFor
\AlgPart{Commit}
\State \textbf{if} no $c_k$ is a valid video \textbf{then return} $Y_t \gets Y_{t-1}$
\State $c \gets c_{k^\star}$, \ $k^\star = \arg\max_j \bigl(\Call{Gate}{z_j, s_e},\, \kappa(z_j)\bigr)$
\State $\mathcal{T} \gets \mathcal{T} \cup \{n\}$, \ $n = \bigl(c,\ e,\ d(e)+\lceil |A|/K\rceil,\ s_e \oplus z_{k^\star}\bigr)$
\State \Return $Y_t \gets X_n$ at the source frame rate
\end{algorithmic}
\end{algorithm}

%% file: generated/tab_mt_main.tex
\textcolor{black!50}{Source clip} & -- & -- & -- & -- & -- & \textcolor{black!50}{14.54} & -- & \textcolor{black!50}{4.08} & -- & \textcolor{black!50}{0.461} & \textcolor{black!50}{5.10} & \textcolor{black!50}{0.756} & -- \\
\midrule
\multicolumn{14}{@{}l}{\emph{Final turn}} \\
Kiwi-Edit & 43.2 & 24.3 & 38.3 & 35.3 & 55.9 & 10.57 & $-19\%$ & 0.72 & $-82\%$ & 0.341 & 4.30 & 0.731 & 144 \\
CoinVE-Edit & 53.5 & 42.2 & 56.1 & 50.6 & 60.3 & 11.47 & $-16\%$ & 0.98 & $-76\%$ & 0.313 & 4.42 & 0.762 & 144 \\
MiniMax H3 & 62.9 & 66.5 & 61.9 & 63.8 & 58.4 & 12.04 & $-11\%$ & 3.09 & $-23\%$ & 0.349 & 4.31 & \textbf{0.768} & 144 \\
\multicolumn{14}{@{}l}{\agentname{} (ours)} \\
\quad VLM: Qwen & 83.3 & 84.6 & \textbf{84.7} & 84.2 & 75.7 & \textbf{16.72} & $\mathbf{+22\%}$ & \textbf{3.99} & $\mathbf{-1\%}$ & \textbf{0.400} & 5.19 & 0.761 & 189 \\
\quad VLM: Gemini & \textbf{83.8} & \textbf{87.7} & 84.5 & \textbf{85.3} & \textbf{76.2} & 16.67 & $\mathbf{+22\%}$ & \textbf{3.99} & $\mathbf{-1\%}$ & 0.385 & \textbf{5.21} & 0.767 & 179 \\
\midrule
\multicolumn{14}{@{}l}{\emph{All turns}} \\
Kiwi-Edit & 49.0 & 42.5 & 48.4 & 46.7 & 69.3 & 13.21 & $-3\%$ & 2.20 & $-46\%$ & 0.368 & 4.62 & 0.745 & 907 \\
CoinVE-Edit & 60.7 & 60.2 & 64.3 & 61.7 & 77.2 & 14.42 & $+5\%$ & 2.70 & $-34\%$ & 0.369 & 4.77 & 0.764 & 907 \\
MiniMax H3 & 65.0 & 75.8 & 64.0 & 68.3 & 67.7 & 13.74 & $0\%$ & 3.69 & $-9\%$ & 0.381 & 4.46 & \textbf{0.775} & 907 \\
\multicolumn{14}{@{}l}{\agentname{} (ours)} \\
\quad VLM: Qwen & 84.2 & 88.1 & 86.4 & 86.2 & 83.1 & 16.26 & $+17\%$ & 4.05 & $\mathbf{0\%}$ & \textbf{0.416} & 5.26 & 0.764 & 1254 \\
\quad VLM: Gemini & \textbf{85.0} & \textbf{88.4} & \textbf{86.7} & \textbf{86.7} & \textbf{84.2} & \textbf{16.37} & $\mathbf{+18\%}$ & \textbf{4.06} & $\mathbf{0\%}$ & 0.405 & \textbf{5.27} & 0.768 & 1166 \\

%% file: sections/experiments.tex
\section{Experiments}
\label{sec:experiments}

\subsection{Setup}
\label{sec:exp-setup}

\noindent\textbf{Conversations.}
We use all \nSessions{} sessions of the benchmark (Sec.~\ref{sec:benchmark}), on 720p clips of $W=49$ frames.

\noindent\textbf{Implementation.}
\agentname{} calls CoinVE-Edit~\cite{coinve200k} with at most two attempts per turn. The refiner and
writer use gemini-3.6-flash, the verifier gemini-3.7-flash (App.~\ref{app:exp-details}). We set $D=3$, $K=5$, and $K_{\mathrm{merge}}=2$
(App.~\ref{app:dev-data}).

\noindent\textbf{Baselines.}
Kiwi-Edit~\cite{kiwiedit}, CoinVE-Edit~\cite{coinve200k}, and MiniMax H3~\cite{minimaxh3} are used
turn by turn: each message edits the previous turn's output, which is how a
single-request editor is used interactively. We call this setting Chain. The editors receive each
message as the user wrote it. Sec.~\ref{sec:exp-ablation} also edits the previous output with
\agentname{}'s instructions. We compare under a fixed editor: Memory-V2V~\cite{memoryv2v} fine-tunes the
video model, and \cite{deepmind2026geminiomniflash} is a closed model, so neither isolates how a given
editor is used.

\noindent\textbf{Metrics.}
Every turn is scored with the checklist of Sec.~\ref{sec:bench-eval} by the CoinVE-Bench evaluator
(gemini-3.6-flash; App.~\ref{sec:bench-clips}). Video quality is measured on every turn, following
CoVEBench~\cite{covebench}, with technical quality (TQ, DOVER++~\cite{dover}), VisualQuality-R1
(VQR)~\cite{visualqualityr1}, physical commonsense (PC, VideoCon-Physics~\cite{videophy}),
aesthetic score (AES~\cite{aesv25}), and motion smoothness (MSM), which can favor suppressed motion~\cite{li2026motionreward}.

\subsection{Main Results}
\label{sec:exp-main}
\begin{figure*}[t]
\centering
\includegraphics[width=\textwidth]{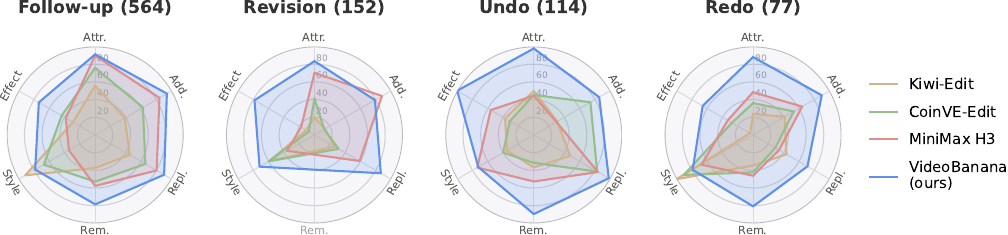}
\caption{Editing accuracy by action and edit type (turns in parentheses). \agentname{} leads on every
action, most clearly on undo.}
\label{fig:mt-ops-radar}
\end{figure*}

Table~\ref{tab:mt-main} compares the systems. With the same editor, \agentname{} raises final-turn
editing accuracy from \nFinalIFCoinve{} to \nFinalIFAgent{} and content preservation from
\nFinalCPCoinve{} to \nFinalCPAgent{} over CoinVE-Edit turn by turn, and is above MiniMax H3,
the strongest baseline (\nFinalIFHthree{} and \nFinalCPHthree{}; App.~\ref{app:h3-view}). Its final outputs are the best on four of the five video-quality metrics and level with MiniMax H3 on the fifth. Their VQR is only
$\nFinalDVQRabsAgent\%$ below the source, against $\nFinalDVQRabsCoinve\%$ for CoinVE-Edit turn by turn.
TQ can exceed the source because it penalizes the blur and noise of real footage
(App.~\ref{app:mt-vq}).
\agentname{} makes \nCallRatio{} times the editor calls of turn-by-turn
editing (App.~\ref{app:agent-runs}).
The gain holds with another VLM or editor (Sec.~\ref{sec:exp-sensitivity}).

On the \nRevertQ{} questions about elements returned to their original state,
\agentname{} answers \nRevertOKAgent{} as expected and CoinVE-Edit turn by turn \nRevertOKCoinve{}. The largest gap is in scope accuracy (\nFinalSPAAgent{}
against \nFinalSPACoinve{} at the final turn), which asks if each edit is clean and confined to its target.

\subsection{Results by Multi-Turn Action and Edit Type}
\label{sec:exp-ops}

\begin{figure*}[t]
\centering
\includegraphics[width=\textwidth]{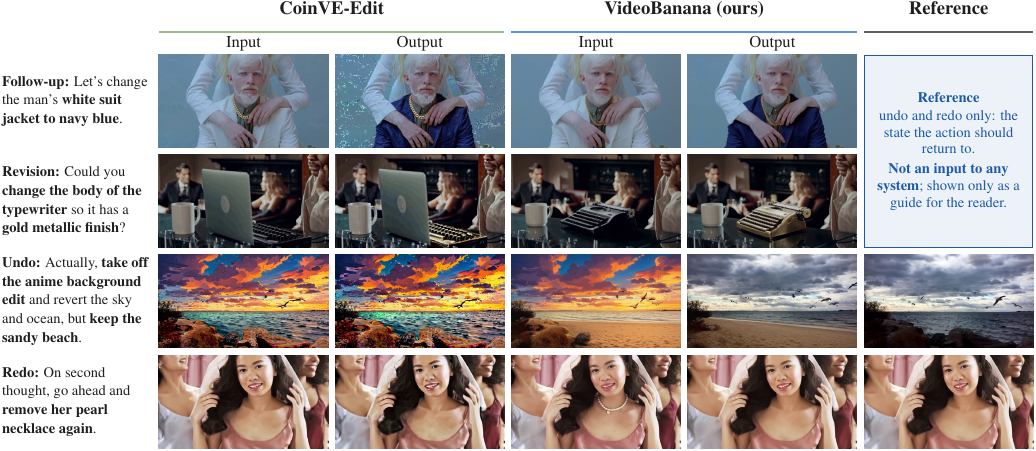}
\caption{Qualitative comparison of \agentname{} with its own editor, CoinVE-Edit, used turn by turn.
Each input is the system's own previous output. Follow-up: CoinVE-Edit's output loses quality relative
to \agentname{}'s, because it re-edits its own output at every turn. Revision: CoinVE-Edit cannot repair
its earlier failed replacement, so the error carries over. Undo: unlike \agentname{}, CoinVE-Edit keeps
no memory of earlier states, so it cannot restore the scene before the undone edit.
Redo: CoinVE-Edit's earlier undo never brought the necklace back, so CoinVE-Edit meets the redo only by
accident; \agentname{} recovers the earlier state from its checkpoints.}
\label{fig:mt-operations}
\end{figure*}

Figure~\ref{fig:mt-ops-radar} scores each turn only on the questions it changes (protocol in
Appendix~\ref{app:mt-ops-types}). Figures~\ref{fig:mt-operations} and~\ref{fig:teaser} show examples, and
App.~\ref{app:agent-failures} shows failure cases.

The turn-by-turn baselines can only edit their latest output, so actions that refer back are hard for
them. A revision fails if the earlier edit did not hold, and an undo must recreate overwritten content. \agentname{} is ahead of every
baseline on every action that refers back. On undos it scores \nOpUndoAllAgent{}, against
\nOpUndoAllBaseLo{} to \nOpUndoAllBaseHi{} for the baselines, and on revisions \nOpRevAgent{}, against
\nOpRevBaseLo{} to \nOpRevBaseHi{}. The baselines' redo scores come mostly from undos that failed
and left the edit in place. Where the undo held, they redo at \nRedoIFHeldBaseMin{} to
\nRedoIFHeldBaseMax{}, against \nRedoIFHeldAgent{} for \agentname{}, whose undo failed before only
\nRedoFailedTurnsAgent{} of the \nBhRedo{} redos (App.~\ref{app:mt-ops-types}). On follow-ups, \agentname{} is above its own
editor on every edit type. Local style is the only follow-up type on which a baseline is higher (Kiwi-Edit,
\nTypeMadeStyleKiwi{}), and there \agentname{} is bounded by its editor, CoinVE-Edit.

\subsection{Results across Turns}
\label{sec:exp-quality}

\begin{figure*}[t]
\centering
\begin{tikzpicture}
\begin{groupplot}[group style={group size=3 by 1, horizontal sep=1.25cm}, mtpanel]
\nextgroupplot[title={(a) Editing accuracy at each turn}, xlabel={Turn}, ylabel={Editing accuracy},
  xmin=0.6, xmax=8.4, ymin=20, ymax=100, xtick={1,...,8},
  legend to name=mtlegend, legend columns=4, legend style={font=\small, draw=none, column sep=8pt}]
\addplot[mtkiwi] table[x=turn, y=Kiwi] {generated/fig_mt_turns_if.dat}; \addlegendentry{Kiwi-Edit}
\addplot[mtcoinve] table[x=turn, y=Coinve] {generated/fig_mt_turns_if.dat}; \addlegendentry{CoinVE-Edit}
\addplot[mththree] table[x=turn, y=Hthree] {generated/fig_mt_turns_if.dat}; \addlegendentry{MiniMax H3}
\addplot[mtagent] table[x=turn, y=Agent] {generated/fig_mt_turns_if.dat}; \addlegendentry{\agentname{} (ours)}
\nextgroupplot[title={(b) Content preservation at each turn}, xlabel={Turn}, ylabel={Content preservation},
  xmin=0.6, xmax=8.4, ymin=20, ymax=100, xtick={1,...,8}]
\addplot[mtkiwi] table[x=turn, y=Kiwi] {generated/fig_mt_turns_cp.dat};
\addplot[mtcoinve] table[x=turn, y=Coinve] {generated/fig_mt_turns_cp.dat};
\addplot[mththree] table[x=turn, y=Hthree] {generated/fig_mt_turns_cp.dat};
\addplot[mtagent] table[x=turn, y=Agent] {generated/fig_mt_turns_cp.dat};
\nextgroupplot[title={(c) VQR change from the source}, xlabel={Turn}, ylabel={$\Delta$VQR (\%)},
  xmin=0.6, xmax=8.4, ymin=-100, ymax=10, xtick={1,...,8}]
\draw[cNeutral, dashed] (axis cs:0.6,0) -- (axis cs:8.4,0);
\addplot[mtkiwi] table[x=turn, y=Kiwi] {generated/fig_mt_turns_dvqr.dat};
\addplot[mtcoinve] table[x=turn, y=Coinve] {generated/fig_mt_turns_dvqr.dat};
\addplot[mththree] table[x=turn, y=Hthree] {generated/fig_mt_turns_dvqr.dat};
\addplot[mtagent] table[x=turn, y=Agent] {generated/fig_mt_turns_dvqr.dat};
\end{groupplot}
\node[anchor=south] at ($(group c2r1.north)+(0,0.45cm)$) {\pgfplotslegendfromname{mtlegend}};
\end{tikzpicture}
\caption{Editing accuracy, content preservation, and video quality across the turns of a session. The
baselines decline on all three as their outputs undergo successive generations. \agentname{} keeps its
outputs within a few generations of the source and declines far less.}
\label{fig:mt-turns}
\end{figure*}
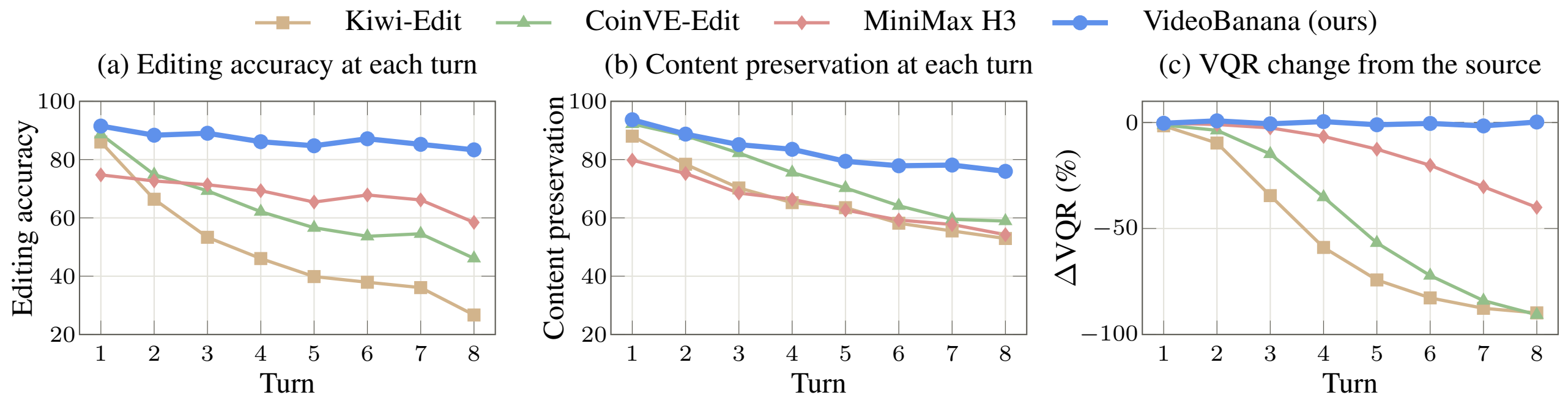

Figure~\ref{fig:mt-turns} follows the systems over the turns. (a) At the first turn, where every
system edits the source clip, CoinVE-Edit turn by turn is within \nTurnOneGapCoinve{} points of 
\agentname{}. From then on the dedicated editors lose editing accuracy as their outputs are edited again
and again (CoinVE-Edit \nTurnIFCoinveFive{} at turn 5), and MiniMax H3 declines more slowly. \agentname{}
stays between \nTurnIFAgentMin{} and \nTurnIFAgentMax{} at every turn, because it rebuilds from earlier
checkpoints: no output of \agentname{} has gone through more than \nBhMaxDepth{} generations. (b) Content
preservation falls for every system as edits accumulate, but more slowly for \agentname{}. (c) Video quality follows the same
pattern: the dedicated editors' VQR is almost unchanged for one or two generations and then falls
quickly (CoinVE-Edit $\nTurnDVQRabsCoinveFive\%$ below the source at turn 5), while \agentname{} stays
within $\nTurnDVQRAgentMaxAbs\%$ of the source at every turn (App.~\ref{app:mt-vq}).

\subsection{Ablation Study}
\label{sec:exp-ablation}

\begin{table}[t]
\centering
\small
\caption{Gain from each part of \agentname{} on its active turns, with everything else held fixed. A
turn is active for a part if switching the part off changes its output.}
\label{tab:mt-ablation}
\setlength{\tabcolsep}{0pt}
\begin{tabular*}{\columnwidth}{@{\extracolsep{\fill}}lrccccc@{}}
\toprule
& Turns & SA & SPA & EP & EA & CP \\
\midrule
\input{generated/tab_mt_ablation.tex} 
\bottomrule
\end{tabular*}
\end{table}

The three parts of \agentname{} depend on one another. Removing one alone leaves the others without their
input, so a cumulative ablation changes several parts in each row (App.~\ref{app:controlled}). We
instead isolate each part with everything else held fixed
(Table~\ref{tab:mt-ablation}). Each comparison changes one decision of a turn: where it starts, what it
executes, or whether its sample is accepted. A part is measured on its \emph{active} turns, where
switching it off changes the output. Elsewhere the two sides are the same and would only dilute the gain.

\noindent\textbf{Entry selection.}
Entry selection is active where the goal sends \agentname{} back to an earlier checkpoint. Either the
previous output shows an unwanted edit, or an earlier checkpoint already shows the goal. On these turns
we also edit \agentname{}'s previous output, with the same goal and the same instruction writer. Both sides use one sample and no verifier. The selected entry still gives clearly higher semantic accuracy and content preservation. In
\nOneStepReuseTurns{} of these turns, all undos or redos, the checkpoint is returned without an editor
call, so the user gets back exactly the video seen before. Editing the previous output loses most on
these reused turns: most of them take out an earlier edit, and the editor often leaves part of it. On
the turns that edit again from an earlier checkpoint, both sides reach about the same editing accuracy,
and the selected entry preserves more content (App.~\ref{app:controlled}). Over a
session, it also keeps every output within \nBhMaxDepth{} generations of the source
(Sec.~\ref{sec:exp-quality}).

\noindent\textbf{Goal tracking.}
Goal tracking is active where the conversation has ended an edit, by an undo or a revision. Only then
does the goal differ from the list of messages. At the final turn of these sessions, we execute either
all of the user's messages or the tracked goal, each once on the source. Of the \nCtlEndedQ{} questions
about undone or replaced edits,
$\nCtlEndedPctOff\%$ are answered as expected without it and $\nCtlEndedPctOn\%$ with it.

\noindent\textbf{Verifier.}
The verifier is active where it rejects the first sample and the second one is committed. On these turns
we compare the two samples. The entry is the same for both. Rejecting and resampling raises every part
of EA, and CP by less. Most rejections are for an edit that was not
made ($\nGateEditPct\%$), and fewer for damage to the rest of the video ($\nGateRestPct\%$;
App.~\ref{app:gate-reasons}). Over all sessions, committing the first sample at the final turn
lowers its EA from \nFinalIFAgent{} to \nVerifAllFinalIFOff{}. Resampling also helps without a tree: on the \nStCases{} single
requests of CoinVE-Bench, it raises CoinVE-Edit's average score from \nStAvgCoinve{} to \nStAvgAgent{}
(App.~\ref{app:single-turn}).

\subsection{Sensitivity Analysis}
\label{sec:exp-sensitivity}

We test whether the results depend on one setting or on one model. We vary the depth limit of entry
selection, replace the VLM with an open model, and use a weaker video editor.

\noindent\textbf{Depth limit.}
We test the edge case of the depth limit $D$ (Sec.~\ref{sec:tree}), $D=1$, where
every turn returns a stored output or edits the source (App.~\ref{app:depth-limit}). EA does not separate
it from the default $D=3$: about half the sessions tie, and the rest split both ways. With $D=1$, the
edits kept from earlier turns are regenerated at every turn and change between turns, which the
checklist does not see. With $D=3$, a turn can continue from an output the user has seen, which improves
consistency: consecutive outputs have a higher SSIM.

\noindent\textbf{VLM.}
By default, \agentname{} uses Gemini models, the family of the benchmark's evaluator
(Sec.~\ref{sec:bench-eval}). To test whether the results depend on this family, we replace the refiner,
the instruction writer, and the verifier with Qwen3.8-27B, an open model that runs locally. No model of
\agentname{} then shares the evaluator's family. Its final-turn EA stays within \nQwenAllIFGapAbs{} points
of the Gemini version (\nQwenAllIFQwen{} against \nQwenAllIFAgent{}; App.~\ref{app:open-vlm}), so
it does not need a closed VLM.

\noindent\textbf{Video editor.}
To test if the gain depends on the editor, we replace CoinVE-Edit with Kiwi-Edit, the weakest editor
of Table~\ref{tab:mt-main}, not trained for compositional instructions, and keep the open
VLM. Even in this weak configuration, \agentname{} raises Kiwi-Edit's final-turn EA from \nOEdFinalIFKiwiTbt{} to
\nOEdFinalIFKiwiAgent{} and CP from \nOEdFinalCPKiwiTbt{} to \nOEdFinalCPKiwiAgent{}
(App.~\ref{app:other-editor}). The EA gain is as large as with CoinVE-Edit, so it is not tied to one
editor.

%% file: generated/tab_mt_ablation.tex
Entry selection & 141 & +10.4 & $-$0.4 & +1.1 & +3.7 & +5.7 \\
Goal tracking & 72 & +11.5 & +1.4 & +7.5 & +6.8 & +2.8 \\
Verifier & 252 & +16.1 & +13.7 & +16.7 & +15.6 & +3.1 \\

%% file: sections/conclusion.tex
\section{Conclusion}
\label{sec:conclusion}

We introduced \benchname{}, a benchmark for multi-turn, conversational video editing. On it,
single-request editors used turn by turn lose quality and fail to restore most of what
an undo asks for. We then presented \agentname{}, which tracks the goal of the conversation over a tree
of verifier-scored checkpoints. With the same editor, it is more accurate, undoes reliably, and largely keeps video
quality close to the source. Requests relative to a generated output, such as ``a little brighter'',
are a natural next step: every output is already a checkpoint they can refer to, and slider-based image editors control how strong an edit is~\cite{zarei2026slidereditpp}. We hope \benchname{} and \agentname{}
help move video editing from single requests to conversations.

%% file: sections/appendix.tex
\noindent\textbf{Appendix outline.}
\begin{tightitemize}
    \item \emph{\ref{app:bench-compare-full} Additional Related Work.} Video editors and their evaluation,
          multi-turn image editing benchmarks and methods, agents, and dialogue state tracking.
    \item \emph{\ref{app:benchmark} Benchmark Construction.} More sessions, scope, statistics, source clips, the
          formal actions, the construction steps, the per-turn checklists, and the review.
    \item \emph{\ref{app:agent-details} Details of \agentname{}.} Goal tracking, entry selection, the
          verification gate, and the single-turn editing agent.
    \item \emph{\ref{app:results} Additional Results.} Experimental details, development and tuning, results with
          other VLMs and editors, run statistics, and results on single-request editing, by action and edit
          type, and by turn; the controlled comparisons, the verifier's rejections, and the depth limit.
    \item \emph{\ref{app:qualitative} Qualitative Results.} Four sessions turn by turn, two failures of the baselines, and failure
          cases of \agentname{}.
    \item \emph{\ref{app:prompts} Prompt Templates.} The prompts of all model calls.
\end{tightitemize}

\section{Additional Related Work}
\label{app:bench-compare-full}

\begin{table}[t]
\centering
\small
\caption{Comparison with multi-turn image editing benchmarks, with the video benchmark \benchname{} in
the last row. Columns as in Table~\ref{tab:bench-compare}. \pmark: only partly.}
\label{tab:bench-compare-full}
\setlength{\tabcolsep}{1pt}
\newcommand{\twoline}[2]{\begin{tabular}[b]{@{}c@{}}#1\\#2\end{tabular}}
\begin{tabular*}{\columnwidth}{@{\extracolsep{\fill}}lccccccc@{}}
\toprule
& & \multicolumn{2}{c}{Turns} & & & & \\
\cmidrule(lr){3-4}
Benchmark & \twoline{Ses-}{sions} & \twoline{per}{session} & total & \twoline{Conver-}{sational} & Undo & \twoline{Revi-}{sion} & \twoline{State-}{based} \\
\midrule
EdiVal-Bench~\cite{edival} & 572 & 3 & 1{,}716 & \xmark & \xmark & \xmark & \xmark \\
LME-Bench~\cite{lmebench} & 100 & 10 & 1{,}000 & \xmark & \xmark & \xmark & \xmark \\
MICE-Bench~\cite{editr2} & 720 & 3 & 2{,}160 & \cmark & \xmark & \xmark & \pmark \\
HDD-Bench~\cite{agentbanana} & 96 & 3 & 288 & \cmark & \cmark & \xmark & \cmark \\
\midrule
\textbf{\benchname{}} & 144 & 5--8 & 907 & \cmark & \cmark & \cmark & \cmark \\
\bottomrule
\end{tabular*}
\end{table}

Beyond training on instruction data, video editors unify editing with generation~\cite{vace,vino,omniweaving},
take reference images~\cite{kiwiedit}, or separate semantic anchoring from motion~\cite{sama}. Industrial
omni-modal generators also edit video~\cite{seedance2,klingomni,klingo3,minimaxh3}, and video-audio models generate and edit both modalities jointly~\cite{sharma2026videoaudio}. Video editing
benchmarks increasingly score requests through questions answered by a VLM~\cite{tifa,dsg}, and fine-grained MLLM judges agree closely with human ratings of image edits~\cite{liu2026mllmjudges}.

Multi-turn editing has been benchmarked mostly on images, in sessions of three~\cite{magicbrush,edival}
to ten~\cite{lmebench} turns in which every turn is a fresh request on the previous output
(Table~\ref{tab:bench-compare-full}). A few image benchmarks contain conversation-dependent turns: content
memory, pronouns, and version backtracking in ImgEdit-Bench~\cite{imgedit} and MICE-Bench~\cite{editr2},
and a target scene state with undo in HDD-Bench~\cite{agentbanana}, all in three-turn sessions. None of
them has revision turns. For video, WBench~\cite{wbench} evaluates multi-turn interaction with world
models, not the editing of a given clip.

Multi-turn image editing methods track the request over the dialogue, treating undo and redo as
intents~\cite{lin2018dialogedit}, using a finite-state machine~\cite{jiang2021talktoedit}, or
re-editing the initial image to limit error accumulation~\cite{cui2023chatedit}. Others condition on
the first image and recent turns~\cite{anchoredit} or on the whole session~\cite{qu2026vincie}.
Non-Markov editing~\cite{zhang2026nonmarkov} trains a history-conditioned model on rollbacks and
revisions. Like the video methods in Sec.~\ref{sec:related}, they change or condition the editor.

Before LLM agents, language-driven image editors mapped a request to editing operations and their parameters~\cite{shi2020languagedriven,shi2021planning}.
Agents decompose a request, call generators or editing tools, and check the result, for image
generation and editing~\cite{wang2024genartist,wang2024compagent,agentbanana,gupta2025costa,mukherjee2026agenticstyling} and for
video generation and
editing~\cite{wen2026avecompass,wu2026mmlve,tu2024spagent,liang2025univa,wang2024lave,sandovalcastaneda2025editduet}.
Restating the goal each turn is a form of dialogue state
tracking~\cite{williams2013dstc,dernoncourt2017robustdst,budzianowski2018multiwoz}. Unlike fixed slots, our intents are
free-form states of video elements, and a verifier checks each of them on the video.

\FloatBarrier

\section{Benchmark Construction}
\label{app:benchmark}

\renewcommand{\dblfloatpagefraction}{0.9}
\begin{figure*}[p]
\centering
\includegraphics[width=\textwidth]{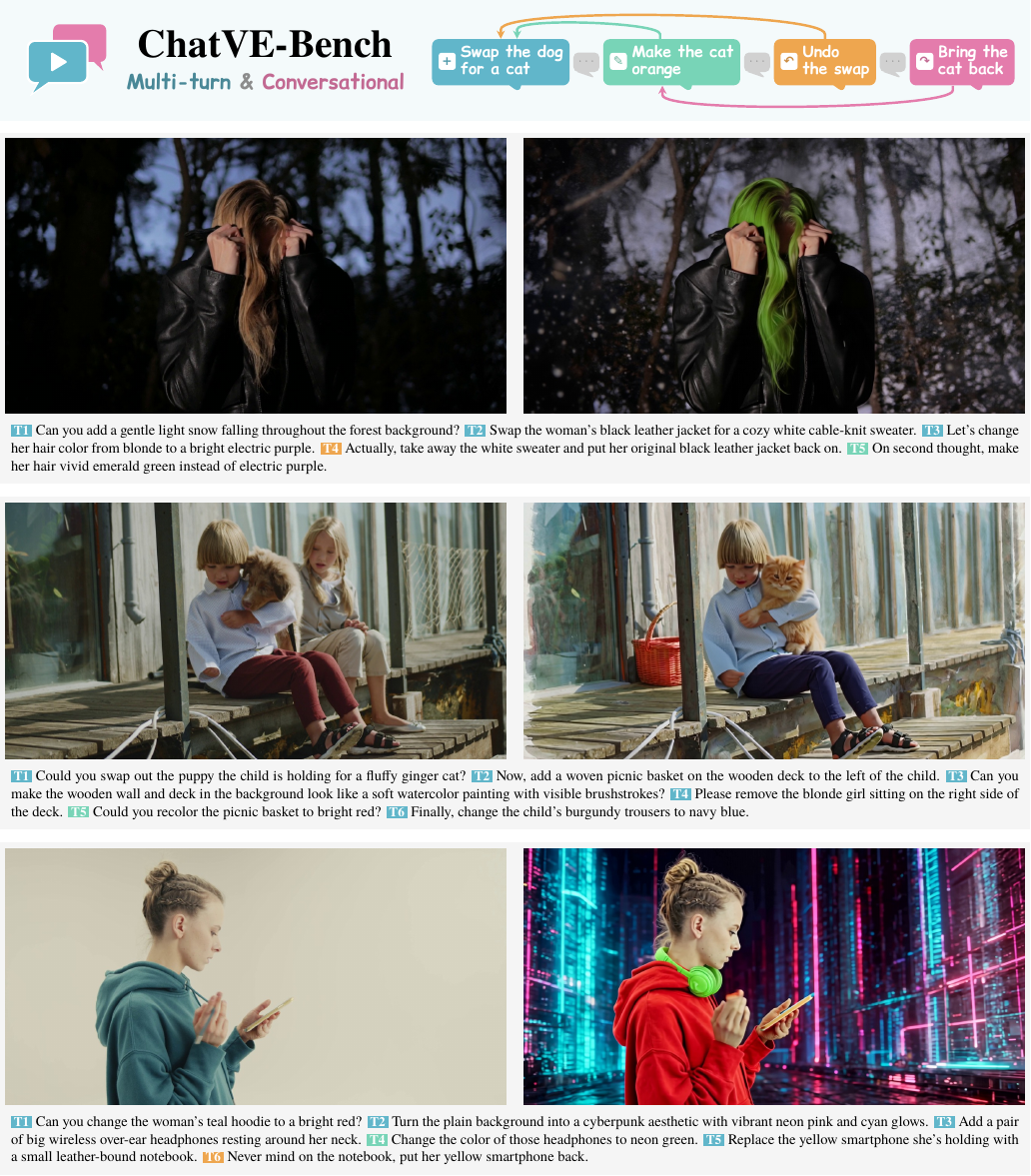}
\caption{Examples from \benchname{} with \agentname{}'s results. Left: the source video. Right:
\agentname{}'s output after the last turn. Below: the requests of all turns of the session, in order, each led
by its turn (Tk).}
\label{fig:bench-showcase}
\end{figure*}

\begin{figure*}[p]
\centering
\includegraphics[width=\textwidth]{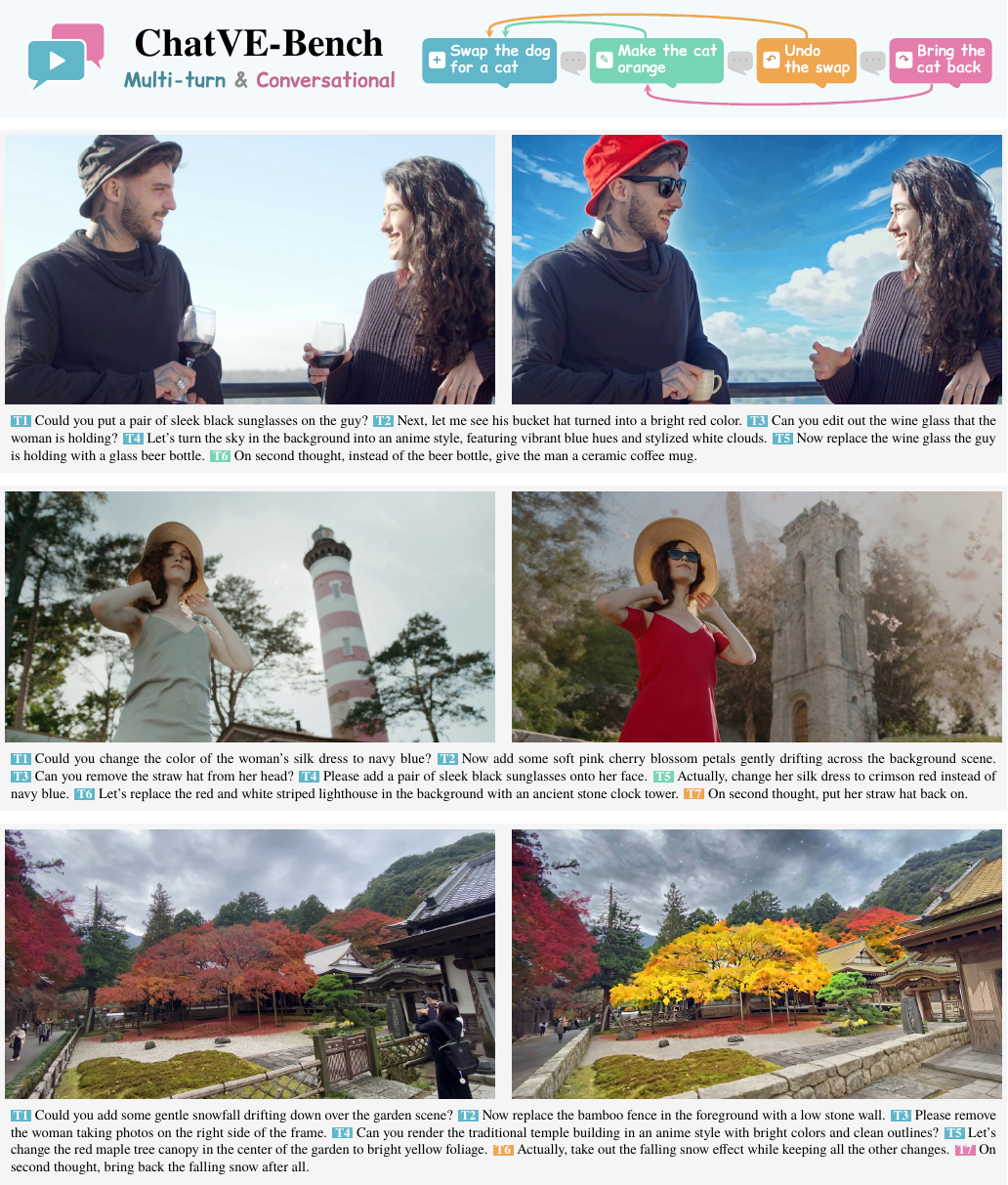}
\caption{More examples from \benchname{} with \agentname{}'s results, as in Figure~\ref{fig:bench-showcase}.}
\label{fig:bench-showcase-2}
\end{figure*}

Figures~\ref{fig:bench-showcase} and~\ref{fig:bench-showcase-2} show six sessions of \benchname{} besides
the three of Figure~\ref{fig:bench-examples} in the main text.

\subsection{Scope}
\label{app:bench-scope}

\benchname{} measures one ability: following a conversation over several turns. Everything else is held at a
level that current instruction-guided video editors already handle, so that a result reflects this ability
and no other.
\begin{tightitemize}
    \item \emph{Clip length.} Every clip has 49 frames, about 2\,s, except one of 41 frames whose source video is shorter (App.~\ref{sec:bench-clips}). An editor for
          longer videos can still be evaluated on a short clip, but an editor for short clips cannot edit a
          long video. Short clips therefore admit the most editors. A turn then succeeds or fails on the
          request itself, not on the length of the video.
    \item \emph{Clip content.} A main subject is in view in every frame, and every clip has at least two
          editable elements besides the background. Each request thus has a target that stays visible, and
          a session can edit distinct elements.
    \item \emph{Edits.} Every edit is one atomic change of appearance, of the six types of
          Sec.~\ref{sec:bench-taxonomy}. Camera motion, object motion, and timing are not edited. Current
          editors support appearance edits. Such an edit also sets a property of an element to a value,
          which revisions, undos, and redos act on.
    \item \emph{One request per turn.} Each turn makes one request, with one action and one edit type.
          Several edits in one request are another ability, compositional editing, which
          CoinVE-Bench~\cite{coinve200k} measures. With one request per turn, every
          result can be read by action and by edit type (Sec.~\ref{sec:exp-ops}).
    \item \emph{Session length.} A session has 5 to 8 turns. This is long enough for a request to revise,
          undo, or redo an edit made several turns earlier, and for the effect of repeated editing to show
          (Fig.~\ref{fig:mt-turns} and App.~\ref{app:mt-vq}).
    \item \emph{User messages.} An LLM writes the messages from the blueprint, and we review them
          (App.~\ref{sec:bench-review}). The expected state of every turn is therefore known exactly and
          does not depend on how a message is read.
\end{tightitemize}

\subsection{Statistics}
\label{app:bench-stats}

\begin{table*}[t]
\centering
\small
\caption{Core question templates (Semantic Accuracy). $X$: target; $c$: new value; $A$/$B$: the
original and the new object.}
\label{tab:bench-core}
\begin{tabular}{@{}l@{\hspace{1.5em}}l@{\hspace{1.5em}}l@{}}
\toprule
Edit type & Active & Reverted \\
\midrule
Attribute, style & AB-MCQ on the color (material, pattern, style) of $X$: original vs.\ $c$
                 & $X$ as in the source (Dual-TF); never $c$ \\
Addition, effect & $B$ is seen at its location & $B$ is never seen \\
Replacement & $A$ is never seen; $B$ is seen at $A$'s location & $A$ as in the source (Dual-TF); $B$ is never seen \\
Removal & $A$ is never seen & $A$ as in the source (Dual-TF) \\
\bottomrule
\end{tabular}
\end{table*}

Figure~\ref{fig:bench-overview} and Table~\ref{tab:bench-compare} give the turns by action and edit
type and the size of the checklists. In addition:
\begin{tightitemize}
    \item 46, 36, 35, and 27 sessions have 5, 6, 7, and 8 turns, with between 2 and 5 follow-ups in each session;
    \item \nOpTurnsUndoLast{} undos are of the last step and \nOpTurnsUndoEarlier{} of an earlier step;
    \item the checklists contain \nDistinctQuestions{} distinct questions
          (\nDistinctQuestionsPerSession{} per session); the final-turn ones contain
          \nFinalQuestions{} questions (\nFinalQuestionsPerSession{} per session): SA 1{,}048,
          EP 515, SPA 325, and CP 413.
\end{tightitemize}

\noindent\textbf{Question types and their counts.}
The checklists use the four question types of CoinVE-Bench~\cite{coinve200k}. Single-TF is a yes/no
question on the output (1{,}445 of the final-turn questions), and Dual-TF a yes/no question that
compares the source and the output (259). AB-MCQ chooses between two options on the output (184).
Score-MCQ rates from 0 to 10 how well the output preserves the source (413).

\subsection{Source Clips}
\label{sec:bench-clips}

The source videos are the 300 videos of VEFX-Bench~\cite{vefxbench}.
\begin{tightitemize}
    \item \emph{Clip.} Every system receives the same clip $W$: the first 49 frames after resampling
          to 24\,fps (about 2\,s; 41 for the one source video that is shorter), resized to about 720p with the rule of CoinVE-Edit. Videos that fade
          in from black start at the first stable frame. Forty-nine frames is CoinVE-Edit's native
          input length. Editors with a longer minimum input run on a longer segment that starts at
          $W$ and are cut back to $W$ for evaluation.
    \item \emph{Screening.} A clip is usable if a main subject is visible in every frame and it has
          at least two editable elements besides the background. We flagged the lowest 10\% on five
          VBench and DOVER quality scores~\cite{vbench,dover} and inspected all 300 clips. We keep 121,
          keep 53 borderline ones for a second check during construction, and drop 126, most of them
          landscapes, aerial views, or texture close-ups without an object to edit.
\end{tightitemize}

\subsection{Edit Types and Actions}
\label{app:bench-ops}

\begin{figure}[t]
\centering
\includegraphics[width=\columnwidth]{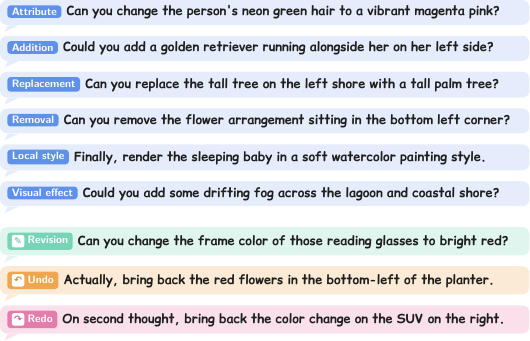}
\caption{A user message from \benchname{} for each edit type and action.}
\label{fig:bench-messages}
\end{figure}

\begin{figure*}[t]
\centering
\includegraphics[width=\textwidth]{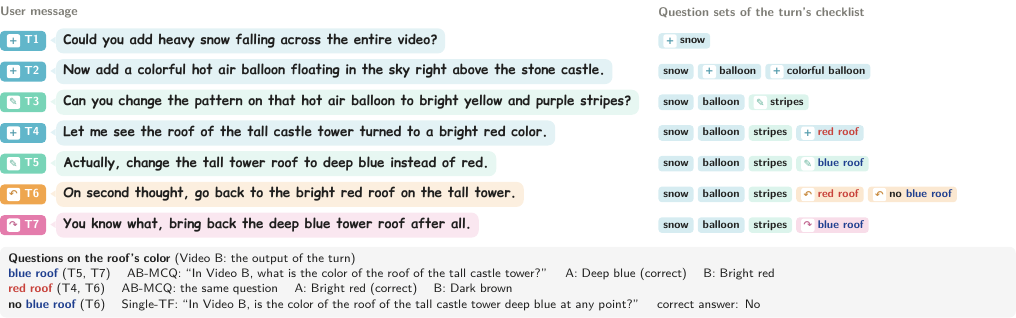}
\caption{Checklist of session 0261 at each turn. Each chip is a question set of one edit, in the color of
the turn that added it. An action icon marks the sets that are new at that turn.}
\label{fig:bench-example}
\end{figure*}

Figure~\ref{fig:bench-messages} gives a user message from the benchmark for each edit type and action
of Sec.~\ref{sec:bench-taxonomy}. The rest of this section defines the edits and actions formally.

\noindent\textbf{Definition of edits and their states.}
An action is defined by how a turn changes the edits of its session, so we first define an edit and its
states. Every turn either makes an edit or refers to an edit made earlier. An edit $e=(o,\pi,v)$ sets
property $\pi$ of element $o$ to value $v$; for an addition, a replacement, or an effect, $\pi$ is
the presence of the content it creates. Let $\mathcal{E}_t$ be the edits made in turns $1,\ldots,t$.
After turn $t$, each $e\in\mathcal{E}_t$ is \emph{active} ($e\in A_t$), \emph{undone}
($e\in D_t$), or \emph{superseded} by a revision of its value; $A_0=D_0=\emptyset$. We write
$\mathrm{new}(e)$ for the elements that $e$ creates (empty for an attribute, style, or removal edit),
$e'\prec e$ if $e$ edits an element in $\mathrm{new}(e')$ ($e$ \emph{builds on} $e'$),
$\mathrm{desc}(e)$ for the active edits that build on $e$ directly or indirectly, and
$\mathrm{rep}(e)$ for the edit whose value $e$ replaced. A request that creates an element
also sets the properties it states (``a black kettlebell'' sets its color); we count each as
an edit of the new element in the same turn, so that a later change of a stated property
replaces its value and a change of a property left open edits the created element. The expected video after turn $t$ is
\begin{equation}
S_t = V_0 \oplus A_t ,
\label{eq:expected-state}
\end{equation}
the source with every active edit applied. It depends only on $A_t$, never on an earlier output, so
an output that failed to show an edit does not change what later turns expect.

\noindent\textbf{Definition of the actions.}
Each turn has exactly one action, given by how it changes the active set. A follow-up refers back to
no earlier edit; the others refer back to an earlier edit $e'$. Whether a turn refers back
depends on what it requests: changing a property that an active edit already
set is a revision even without ``instead''. With $e_t$ the turn's edit:
\begin{tightitemize}
    \item \emph{Follow-up}: $o\in V_0$ and no earlier edit touches $o$.\newline
          $A_t=A_{t-1}\cup\{e_t\}$
    \item \emph{Revision of a created element}: $o\in\mathrm{new}(e')$ for an active $e'$, and no active
          edit sets $\pi$ of $o$; both edits stay.\newline
          $A_t=A_{t-1}\cup\{e_t\}$
    \item \emph{Revision of a value}: $e'=(o,\pi,v)\in A_{t-1}$, $v'\neq v$, and $\mathrm{desc}(e')=\emptyset$ (no active edit builds on $e'$).\newline
          $A_t=(A_{t-1}\setminus\{e'\})\cup\{e_t\}$
    \item \emph{Undo} of any active $e'$, of the \emph{last step} if $e'$ is the edit of turn $t-1$ and
          of an \emph{earlier step} otherwise.\newline
          $A_t=\bigl(A_{t-1}\setminus(\{e'\}\cup\mathrm{desc}(e'))\bigr)\cup\{\mathrm{rep}(e')\}$
    \item \emph{Redo} of an undone $e'\in D_{t-1}$; $F$ holds the edits its undo cascaded to
          (\emph{with dependents}) or none (\emph{parent only}).\newline
          $A_t=\bigl(A_{t-1}\cup\{e'\}\cup F\bigr)\setminus\{\mathrm{rep}(e')\}$
\end{tightitemize}
Here $\mathrm{rep}(e')$ is empty unless $e'$ replaced a value.

\subsection{Construction Steps}
\label{app:bench-steps}

A session is built in four steps. Steps A, C, and D call gemini-3.6-flash on 8 evenly spaced frames
of $W$; step B is a program. A program checks every model output and regenerates it with the error
message until it passes. Every output of steps A, C, and D is reviewed (App.~\ref{sec:bench-review}).

\noindent\textbf{A. Candidate edits.} The model
\begin{tightitemize}
    \item checks the clip again and skips it if the subject is unclear or the frames are blurred or
          occluded;
    \item lists editable elements seen in at least 7 of the 8 frames;
    \item proposes about eight candidate edits in the atomic-edit format, each with a second value
          for a possible revision and, if it creates an element, one further edit of it.
\end{tightitemize}
A program picks 2--5 per clip on distinct, non-overlapping targets, balancing edit
types over the benchmark. As in CoinVE-Bench, only a revision edits an element again.

\noindent\textbf{B. Schedule.} A program shuffles the selected edits and inserts the turns that
refer back to them:
\begin{tightitemize}
    \item With fixed probabilities, it adds a revision of a created element, a revision of a value,
          an undo, and, after an undo, a redo. At least one is added; a session has 5--8 turns.
    \item A revision comes after the edit it revises. An undo can target any earlier step, not only the last one.
    \item When an undo also removed revisions of the element the undone edit created, the program
          draws whether the redo brings them back (\emph{with dependents}) or not (\emph{parent only}).
    \item An undo may not leave the session without any edit, since returning the source video would
          then score fully.
    \item After each insertion, the program replays the expected state of every edit and rejects
          invalid drafts.
\end{tightitemize}
The result, the \emph{blueprint}, holds each turn's action and the expected state after it. It is
the ground truth of the session.

\noindent\textbf{C. Question banks.} Each edit gets a question set for every state the blueprint
puts it in (App.~\ref{app:bench-states}). As a question must hold at every turn it is
used in, questions are of four kinds:
\begin{tightitemize}
    \item \emph{Core} questions decide whether the edit is in the expected state. They come from fixed
          templates in the CoinVE-Bench formats (Table~\ref{tab:bench-core}); the model only fills
          slots from the source clip, such as how to name the target.
    \item \emph{Free} questions: the model adds one or two to the active set, on edit persistence and
          on whether the change stays inside the target. They mention only the edit's own target and
          elements that no edit touches.
    \item \emph{Value} questions keep the checklist exact when a revision changes the look of an
          object an earlier edit created. The earlier edit's questions then name the object without
          that property (``a small glass of liquid''), and the value set asks for the object as first
          named (``a small glass of orange juice'') while the revision is not in effect.
    \item \emph{Preservation} questions (CP, 0--10) are asked at every turn: one
          template question on the shape, position, and movement of an attribute or style edit's
          target, and up to four model questions on elements no edit touches.
\end{tightitemize}

\noindent\textbf{D. User messages.} The model writes one user message per turn from the blueprint and the frames:
\begin{tightitemize}
    \item a follow-up states the change concretely;
    \item a revision refers back (``that balloon'', ``instead'');
    \item an undo names the content of the undone edit, and undoing a revision names the value that the
          target returns to (``go back to the bright red roof'');
    \item a redo never restates the value of a revision of a created element, so a system has to
          remember it.
\end{tightitemize}

\subsection{Per-Turn Checklists}
\label{app:bench-states}

At turn $t$, each edit adds the question set of its state:
\begin{tightitemize}
    \item \emph{In effect} (active or redone): its active set, and for an addition or replacement whose
          later revision of the look is not in effect, also its value set. A redo with
          \emph{parent only} also adds the absent set of each revision it leaves out.
    \item \emph{Superseded by a revision of its value}: the revision's active set, and for an addition,
          replacement, or effect also its replaced set (the superseded content is absent).
    \item \emph{Undone follow-up}: its reverted set (the target looks as in the source, and the edited
          value never appears).
    \item \emph{Undone revision}: its absent set (its value never appears). Undoing a revision of a
          value also brings back the active set of the earlier edit that it replaced.
    \item \emph{Undone by a cascade}: nothing.
\end{tightitemize}
Every turn also asks the session's preservation questions.

\noindent\textbf{Example of a checklist.}
Figure~\ref{fig:bench-example} shows one session. The questions of the snow, the balloon, and its
stripes keep the same text from the turn they are introduced to the end. After the undo of turn 6, the
checklist switches back to the red-roof questions and adds that deep blue must not appear; the redo of
turn 7 switches back to the deep-blue questions.

\subsection{Review of the Benchmark}
\label{sec:bench-review}

Every candidate edit, question bank, and user message is reviewed against its clip, without
looking at any system output. An AI tool checks every item against written criteria:
\begin{tightitemize}
    \item a target is clearly visible, large enough to judge, and singled out unambiguously by the wording;
    \item an edit can be carried out on its clip and matches its edit type (one of the six of Sec.~\ref{sec:bench-taxonomy});
    \item a question is not passed by an unedited video, not failed by a correct edit, and not about
          another edit's element;
    \item a user message expresses its turn's action unambiguously.
\end{tightitemize}
It deletes or fixes failing items; it adds no edits. One of the authors decides every judgment call
(48 referred items, 29 accepted and 19 rejected). Over the 144 sessions, the review deleted 7.8\% of the
candidate edits and 4.9\% of the questions, and reworded 10.3\% of the user messages. CoVEBench keeps
67.2\% of its questions after manual filtering~\cite{covebench}.
A session costs about \$0.10 of API calls. Finally, the author audited 10 random sessions (67 turns,
209 questions). The user messages had no error. One of the 209 questions named the wrong one of several
fences. The audit also led to the value questions of step C (App.~\ref{app:bench-steps}).

\FloatBarrier
\renewcommand{\dblfloatpagefraction}{0.99}

\section{Details of \agentname{}}
\label{app:agent-details}

This section adds to Sec.~\ref{sec:method} what it leaves out.

\subsection{Goal Tracking}
\label{app:goal-entry}

\begin{tightitemize}
    \item \emph{Granularity.} Each intent is one change to one element: one per changed attribute of an
          existing element, or one per added or replaced element with the appearance it is described with
          (``add a blue brick wall'').
    \item \emph{Refiner output.} Each deactivation and reactivation comes with a reason, and an intent that a new intent replaces is deactivated with it. \agentname{} checks that they lead from the previous active list to the stated one and gives each new intent a permanent identifier for later turns.
    \item \emph{Added elements.} A later change to the appearance of an added element is a new intent with
          the updated description, which replaces the old one.
    \item \emph{Intent text.} An intent carries all new information of the request, such as position and
          relations (``golden particles drift above the counter without covering the roses'').
    \item \emph{Preservation constraints.} Requests that only ask to keep something as it is are kept apart from the
          intents.
\end{tightitemize}

\subsection{Entry Selection}
\label{app:entry}

The ledger of node $n$ holds a score $s_n(i)\in\{1,2,3,4\}$ for every intent judged when $n$ was committed
and inherits the other scores from its parent. An intent is realized at $n$ if $s_n(i)\ge\tau$; unknown
scores count as not realized. With $X_t\subseteq G_t$ the exhausted intents, a candidate entry $e$ has
\begin{align*}
A(e) &= \{\, i\in G_t\setminus X_t : i \text{ not realized at } e \,\} && \text{(execute)}\\
P(e) &= \{\, i\in G_t\setminus X_t : i \text{ realized at } e \,\} && \text{(preserve)}\\
U(e) &= \{\, i\notin G_t : i \text{ realized at } e, && \text{(unwanted)}\\
&\qquad\ i \text{ not replaced by any } j\in A(e) \,\}
\end{align*}
An entry is feasible if $U(e)=\emptyset$, and editing it adds $g(e)=\lceil |A(e)|/K\rceil$ generations. Among
feasible entries with $d(e)+g(e)\le D$ and $|A(e)|\le K$, \agentname{} takes
\begin{enumerate}[leftmargin=1.5em, itemsep=1pt, topsep=2pt, parsep=0pt]
    \item an entry with $A(e)=\emptyset$, reused without editing (the shallowest such entry, then the latest);
    \item otherwise, among entries with $|A(e)|\le K_{\mathrm{merge}}$, the shallowest, then the
          fewest instructions, then the latest;
    \item otherwise, among all feasible entries, the fewest instructions, then the shallowest, then the latest.
\end{enumerate}
If no entry satisfies the limits, the feasible entry with $|A(e)|\le K$ and the smallest $d(e)+g(e)$ is
taken, or the one with the smallest $d(e)+g(e)$ among all feasible entries if none has $|A(e)|\le K$. An
attempt is one editor call with at most $K$ instructions, so an edited entry gains one generation; when
$|A(e)|>K$, the writer has to fit the targets into $K$ instructions, or the run stops with an error. This
case did not occur in our runs.

\subsection{Verification and Gate}
\label{app:gate}

\begin{tightitemize}
    \item \emph{Scores.} Items are scored 1 to 4. For an intent, the score says how long its state
          holds: the whole clip 4, more than half 3, half or less 2, absent 1. The pass level is $\tau=3$.
    \item \emph{Damage.} A preserved intent is damaged if its score drops by two levels or more from the
          entry's ledger.
    \item \emph{Attempts.} A turn makes at most two attempts, the second with the next seed, and stops at
          the first pass.
    \item \emph{No pass.} The turn then keeps the candidate with the highest composite score, the
          equal-weight mean of instruction following, preservation, and quality. Each part averages its
          items, and each score $s$ counts as $(s-1)/2$, capped at 1 (that is, at level 3, also with Qwen), so a
          surplus on one item cannot make up for a gap on another. Instruction following covers every intent sent to the editor,
          exhausted ones included. Preservation averages two halves: the earlier edits (1 if not
          damaged, else 0) and the non-target content. Ties are broken by the same mean without the cap,
          each score counting as $(s-1)/3$ and an earlier edit by how far its score dropped. A verdict that cannot be read even after
          the repair leaves the candidate's scores unknown: it fails the gate and counts 0 in the composite.
          If no attempt yields a valid video, the previous output is shown again.
    \item \emph{Exhausted intents.} An executed intent that is not realized in the committed output, or whose
          turn yields no valid video, counts one failure. After two failed turns it is exhausted. It stays in
          the goal but is no longer a gate target or a target of entry selection. It is added to a call as an
          extra instruction only if the call then has at most three; such an extra is verified and counts in
          the composite. Otherwise the verifier does not see it. The rule keeps an edit that the editor cannot make from deciding
          every later turn. Without it, such an edit would fail the gate at every later turn and trigger a
          second sample. No stored output would show the whole goal, so none could be reused. The
          rule applies in \nBhExhaustedTurns{} of the \nTurns{} turns, in \nBhExhaustedSessions{} sessions.
\end{tightitemize}

\subsection{Single-Turn Editing Agent}
\label{app:single-turn-agent}

Each attempt runs a single-turn editing agent on the entry video. Its intent compiler (App.~\ref{app:prompts})
first checks the written instructions: it splits them into atomic intents and must cover every
instruction. If it finds a target ambiguous, or gives no valid answer after two repairs, the attempt makes no edit.
Otherwise the agent sends the written instructions, each once and trimmed, to the editor in one call; the
compiled intents only serve this check. The compiler uses the refiner's model. The output becomes a checkpoint of the tree. In single-request mode (App.~\ref{app:single-turn}), the request is the goal and
the source is the only entry. The verifier checks the result, and a rejected candidate is resampled once.

\FloatBarrier

\section{Additional Results}
\label{app:results}

\subsection{Setup}
\label{app:exp-details}

\begin{tightitemize}
    \item \emph{Hardware.} Every editor call, of \agentname{} and of the baselines, runs on one GPU, an NVIDIA
          RTX PRO 6000 Blackwell (96\,GB) or RTX 6000 Ada (48\,GB). A CoinVE-Edit call takes about 12 and 21
          minutes on them. A MiniMax H3 call takes about 13 minutes on the Blackwell GPU and 34 minutes on the
          Ada GPU, where it fits only with part of the model offloaded to the CPU.
    \item \emph{MiniMax H3.} It receives the 124 frames that start at $W$. In chained runs, its full output is
          the next turn's input.
    \item \emph{Agent settings.} The Gemini calls run at temperature 0.1 with at most 8,192 output tokens; the
          Qwen calls use the model's default sampling with at most 16,384. CoinVE-Edit runs 50 denoising
          steps; an editor call is tried up to three times with the same seed if it crashes.
    \item \emph{Checklist evaluator.} We keep CoinVE-Bench's four system prompts, question types, and
          scoring rules, and send the questions as follows. All questions of one type at one turn go into
          one request, each under an opaque id (i0, i1, \dots), since question ids repeat across a turn's
          question sets. The user's request is not sent, so the judge cannot lean towards the request; the
          preservation prompt keeps its wording about an editing instruction. Each clip is sampled to 8
          frames, at temperature 0. A question without a usable answer is asked again, and one still
          unanswered is left out. Each metric pools its questions over the turns it covers. Every system is
          scored the same way.
    \item \emph{Video-quality metrics.} All five are trained predictors used in earlier benchmarks, not
          prompted judges, and none involves Gemini, so they are independent of the checklist
          evaluator. The $\Delta$ columns of Table~\ref{tab:mt-main} average each output's relative change from
          its source over sessions, so they can differ from the change of the means.
    \item \emph{Goal tracking off.} In this setting of Table~\ref{tab:mt-ablation}, the call receives
          every user message as one instruction, up to eight, while the editor is trained on at most five.
          Merging consecutive messages so that no call has more than five instructions does not raise the
          scores: at the final turn of all \nSessions{} sessions, EA \nAblConcatCapIF{} and
          CP \nAblConcatCapCP{}, against \nAblConcatIF{} and \nAblConcatCP{} with every message. The two
          differ only in the \nAblCapSessions{} sessions with more than five turns.
\end{tightitemize}

\subsection{Development and Tuning}
\label{app:dev-data}

Every setting of \agentname{} was fixed before it ran on any benchmark session. Two development sets
were used.
\begin{tightitemize}
    \item \emph{Development conversations.} Conversations on \nDevClips{} VEFX-Bench clips, made with a
          preliminary session generator, and single-request cases. None is part of \benchname{}. The
          prompts, the verifier's included, and the depth limit were developed on these conversations.
    \item \emph{A development split of the benchmark.} \nQwenDevSessions{} benchmark sessions drawn at random.
          Only the open-VLM version used it, to set its verifier's pass level
          (App.~\ref{app:open-vlm}).
\end{tightitemize}

\noindent\textbf{Setting of $D$, $K$, and $K_{\mathrm{merge}}$.}
\begin{tightitemize}
    \item \emph{$D=3$} was set on the development conversations, from CoinVE-Edit used turn by turn: its VQR loss is
          small for the first two generations and grows from the third. It was not chosen by editing accuracy (App.~\ref{app:depth-limit}).
    \item \emph{$K=5$ and $K_{\mathrm{merge}}=2$} are not tuned. They are the largest and the most common
          number of instructions per sample in the editor's training data.
\end{tightitemize}

\noindent\textbf{Development of the verifier prompt.}
To keep development apart from evaluation, the verifier prompt was written on the development conversations. It scores every active intent
in the same way, as a state of the candidate, and leaves damage to earlier edits to the gate
(Sec.~\ref{sec:verify}).
\begin{tightitemize}
    \item \emph{Pass level.} With Gemini, $\tau=3$ was fixed during this development. With Qwen3.8-27B it is
          $\tau=4$ for the intents; render quality and edit exclusivity keep $\tau=3$. This value was chosen on the development split, and the other \nQwenHeldSessions{} sessions
          were then run once (Table~\ref{tab:qwen-dev-held}).
\end{tightitemize}

\noindent\textbf{Overlap with the benchmark.}
To check whether these conversations affect the results, we score the sessions whose
clips were not used there.
\nDevClipSessions{} of the \nDevClips{} clips of these conversations are also used by benchmark sessions,
with new edits, blueprints, questions, and messages. On the other \nHeldSessions{}
sessions, final-turn EA is \nHeldFinalIFAgent{} for \agentname{} and \nHeldFinalIFCoinve{} for
CoinVE-Edit turn by turn, against \nFinalIFAgent{} and \nFinalIFCoinve{} on all \nSessions{}.

\subsection{Results with Other Models and Run Statistics}
\label{app:main-extra}

\begin{table}[h]
\centering
\small
\caption{\agentname{} with Qwen3.8-27B at the final turn, on the sessions used to choose its setting
(Development) and on the others (Held-out). H3 is MiniMax H3 used turn by turn, and $\Delta$ is Qwen minus
Gemini in EA. The gap is smaller on the held-out sessions.}
\label{tab:qwen-dev-held}
\setlength{\tabcolsep}{4pt}
\begin{tabular}{@{}lrcccccc@{}}
\toprule
& & H3 & \multicolumn{2}{c}{Gemini} & \multicolumn{2}{c}{Qwen} & \\
\cmidrule(lr){3-3}\cmidrule(lr){4-5}\cmidrule(lr){6-7}
Sessions & $n$ & EA & EA & CP & EA & CP & $\Delta$ \\
\midrule
\input{generated/tab_qwen_dev_held.tex} 
\bottomrule
\end{tabular}
\end{table}

\noindent\textbf{Sensitivity to the VLM.}
\phantomsection\label{app:open-vlm}
To test whether the results depend on the Gemini family, which also provides the benchmark's evaluator, we
replace Gemini with Qwen3.8-27B in the intent refiner, the instruction writer, and the verifier (the Qwen
rows of Table~\ref{tab:mt-main}); the intent compiler uses the refiner's model and so runs on Qwen3.8-27B as
well. Everything else is unchanged, except that the pass level
for the intents is $\tau=4$ instead of 3. This setting was chosen on \nQwenDevSessions{} benchmark sessions drawn at random and then run
once on the other \nQwenHeldSessions{} (Table~\ref{tab:qwen-dev-held}). The gap to Gemini is smaller on the
held-out sessions. Every system scores higher there, MiniMax H3 included, so we compare the gaps to
Gemini.

\begin{table}[h]
\centering
\small
\caption{Two video editors on \benchname{}, used in Chain and inside \agentname{}. Both
\agentname{} rows use Qwen3.8-27B as the VLM. Calls: editor calls over all turns.}
\label{tab:other-editor}
\setlength{\tabcolsep}{0pt}
\begin{tabular*}{\columnwidth}{@{\extracolsep{\fill}}llccccr@{}}
\toprule
& & \multicolumn{2}{c}{Final turn} & \multicolumn{2}{c}{All turns} & \\
\cmidrule(lr){3-4}\cmidrule(lr){5-6}
Editor & Use & EA & CP & EA & CP & Calls \\
\midrule
\input{generated/tab_other_editor.tex} 
\bottomrule
\end{tabular*}
\end{table}

\noindent\textbf{Sensitivity to the video editor.}
\phantomsection\label{app:other-editor}
To test whether the gains depend on the editor, we replace CoinVE-Edit with Kiwi-Edit~\cite{kiwiedit}, the
weakest editor in Table~\ref{tab:mt-main}. Kiwi-Edit is not trained for compositional instructions, so the
instructions of one call are joined into one prompt. The VLM is Qwen3.8-27B, set as above
(Table~\ref{tab:other-editor}). The editor still bounds the result: \agentname{} scores lower with Kiwi-Edit than
with CoinVE-Edit, and it draws a second sample in \nOEdSecondKiwi{} of \nTurns{} turns,
against \nOEdSecondCoinve{} with CoinVE-Edit.

\noindent\textbf{Behavior and cost.}
\phantomsection\label{app:agent-runs}
To show how \agentname{} reaches its results and what they cost, we report its decisions and calls in
the main run (Gemini and CoinVE-Edit).
\begin{tightitemize}
    \item \emph{Gate.} Of the turns with an editor call, \nBhGatePass{} passed the gate. A second
          attempt, drawn in \nBhSecond{} turns, passed in \nBhSecondPass{}. \nBhComposite{} turns
          kept the better of two failed attempts.
    \item \emph{Reuse.} \nBhReuseUndoLast{} of the \nBhUndoLast{} undos of the last step and \nBhReuseRedo{} of
          the \nBhRedo{} redos return a checkpoint without calling the editor. An undo of an earlier step
          usually cannot (\nBhNoReuseUndoEarlier{} of \nBhUndoEarlier{} turns), because the later edits must
          stay. \agentname{} re-executes them from a checkpoint without the undone edit. In \nBhReuseInPlace{} of
          these undos, the previous output is kept as it is, because the undone edit had failed there.
    \item \emph{Depth.} \agentname{}'s outputs sit at generation depth 0 (\nBhDepthZero{} turn), 1
          (\nBhDepthOne{}), 2 (\nBhDepthTwo{}), or 3 (\nBhDepthThree{}).
    \item \emph{Calls.} \agentname{} made \nBhCalls{} editor calls, against \nTurns{} for each turn-by-turn
          baseline, and \nVlmCalls{} VLM calls (\nVlmPerTurn{} per turn): \nVlmRefine{} to refine the goal,
          \nVlmWrite{} to write instructions, \nVlmCompile{} in the intent compiler, and \nVlmVerify{} to
          verify.
\end{tightitemize}

\subsection{Results on Single-Request Editing}
\label{app:single-turn}

\begin{table*}[t]
\centering
\small
\caption{Single-request compositional editing on CoinVE-Bench~\cite{coinve200k}. Gray rows ($\dagger$) are quoted from the benchmark's authors. The last two rows are run by us under one protocol, and bold marks the better of them. \agentname{} is higher than its own editor on every accuracy, naturalness, and preservation metric.}
\label{tab:single-turn}
\setlength{\tabcolsep}{4.2pt}
\renewcommand{\arraystretch}{1.08}
\newcommand{\mutedcell}[1]{\textcolor{black!45}{#1}}
\begin{tabular*}{\textwidth}{@{\extracolsep{\fill}}lccccccccccccc@{}}
\toprule
& \multicolumn{3}{c}{Editing accuracy}
& \multicolumn{3}{c}{Physical naturalness}
& \multicolumn{1}{c}{Sem. pres.} & & \multicolumn{4}{c}{Video quality} & \\
\cmidrule(lr){2-4}\cmidrule(lr){5-7}\cmidrule(lr){8-8}\cmidrule(lr){10-13}
Method & SA & SPA & EP & AN & SC & MN & CP & Avg & AQ & TQ & CQ & TS & Calls \\
\midrule
\input{generated/tab_st_main.tex} 
\bottomrule
\end{tabular*}
\end{table*}

We use CoinVE-Bench as released. \agentname{} calls only CoinVE-Edit, so we re-run
CoinVE-Edit on all \nStCases{} cases with the released script and checkpoint and score both
ourselves. Other editors are quoted from their authors.
\begin{tightitemize}
    \item \emph{Columns.} Avg is the mean of editing accuracy, physical naturalness, and semantic
          preservation. For the quoted rows, we compute it from the reported scores. AQ, TQ, CQ, and TS are
          aesthetic, technical, and comprehensive quality and temporal stability. Calls counts editor calls.
    \item \emph{Shared outputs.} \agentname{}'s first call is CoinVE-Edit's call (same instructions, seed 0),
          so both produce the same video whenever \agentname{} keeps its first sample (\nStIdentical{} of
          \nStCases{} cases). Each distinct output is scored once, as the mean of three evaluator runs, so the
          two rows differ only on the cases where \agentname{} acted.
    \item \emph{Re-run.} Our CoinVE-Edit re-run is \nStRerunGap{} points below the reported average
          (\nStAvgCoinve{} vs.\ \nStRepAvgCoinve{}). Both rows share the same generations and evaluation.
    \item \emph{Independence.} The verifier and the evaluator are different Gemini versions with different
          prompts, and \agentname{} never sees the evaluator's questions.
\end{tightitemize}

\begin{table}[t]
\centering
\small
\caption{Results of Table~\ref{tab:single-turn} by the verifier's decision on the first output. Acc.: accepted. Rej.\,$\rightarrow$\,Acc.: rejected, and the resample accepted. Rej.\,$\rightarrow$\,Rej.: both rejected. Each cell gives CoinVE-Edit's score and, in small type, \agentname{}'s change. Most of the gain comes from the cases whose resample is accepted.}
\label{tab:single-turn-split}
\setlength{\tabcolsep}{3pt}
\newcommand{\stup}[1]{{\footnotesize\textcolor{green!45!black}{$+$#1}}}
\newcommand{\stdn}[1]{{\footnotesize\textcolor{red!70!black}{$-$#1}}}
\newcommand{\stzero}{{\footnotesize\textcolor{black!45}{$\pm$0.0}}}
\begin{tabular*}{\columnwidth}{@{\extracolsep{\fill}}lcccc@{}}
\toprule
Verifier & EA & PN & CP & Avg \\
\midrule
\input{generated/tab_st_acted.tex} 
\midrule
\input{generated/tab_st_all.tex} 
\bottomrule
\end{tabular*}
\end{table}

\noindent\textbf{Effect of the verifier.}
To see where the gain on single requests comes from, Table~\ref{tab:single-turn-split} splits the cases by the verifier's decision on CoinVE-Edit's output. The
\nStRejected{} cases it rejects are those CoinVE-Edit does poorly on, so the extra call is spent only where it
is needed. The \nStRpCases{} cases whose resample passes carry about four fifths of the overall gain.

\noindent\textbf{Comparison on the remaining metrics.}
On the four video-quality metrics, \agentname{} and CoinVE-Edit are level (at most \nStVQMaxDiff{} apart), as is physical commonsense (both \nStPCAgent{}).

\subsection{Results by Action and Edit Type}
\label{app:mt-ops-types}

\begin{table}[h]
\centering
\small
\caption{Editing accuracy by action and, below each action, by edit type, with the number of turns in parentheses (the numbers of Fig.~\ref{fig:mt-ops-radar}). Each turn is scored on the questions it changes. The last row is the macro average over the four actions. Scores pool the questions of their turns, so a row need not lie between its parts.}
\label{tab:mt-ops}
\setlength{\tabcolsep}{3pt}
\newcommand{\twoline}[2]{\begin{tabular}[b]{@{}c@{}}#1\\#2\end{tabular}}
\begin{tabular}{@{}lcccc@{}}
\toprule
Turns & \twoline{Kiwi-}{Edit} & \twoline{CoinVE-}{Edit} & \twoline{MiniMax}{H3} & \twoline{\agentname{}}{(ours)} \\
\midrule
\input{generated/tab_mt_ops_types.tex} 
\bottomrule
\end{tabular}
\end{table}

\noindent\textbf{Scoring of each action.}
A turn's checklist repeats the questions of every earlier edit, so its score mixes the action of the
turn with what earlier turns left behind. To measure the action alone, Figure~\ref{fig:mt-ops-radar} and
Table~\ref{tab:mt-ops} score only the question groups whose edit or state is new at that turn:
\begin{tightitemize}
    \item for a follow-up or a revision, its own edit's questions;
    \item for an undo, the reverted or absent questions of the undone edit, and for an undone revision of a
          value also the questions of the value it returns to;
    \item for a redo, the questions of the redone edit.
\end{tightitemize}

\noindent\textbf{Difference to whole-turn scores.}
To show what this scoring changes, we also score each turn on all its questions. Two actions then look
different. For CoinVE-Edit turn by turn, an undo of the
last step scores \nOpWholeUndoLastIFCoinve{} on the whole turn but \nOpUndoLastCoinve{} on the undo: the
earlier edits are still in the video, and only the undone one is not reverted. For \agentname{}, an undo of an
earlier step scores \nOpUndoEarlierAgent{} on the undo but \nOpWholeUndoEarlierIFAgent{} on the whole turn:
the undone edit is gone, but not all of the later edits that it rebuilds still hold.

\noindent\textbf{Gain by edit type.}
To see which types gain most, we compare \agentname{} with CoinVE-Edit turn by turn. The
gain is largest for additions (\nTypeMadeAddAgent{} vs.\
\nTypeMadeAddCoinve{}) and visual effects (\nTypeMadeEffectAgent{} vs.\ \nTypeMadeEffectCoinve{}) and smallest
for local style (\nTypeMadeStyleAgent{} vs.\ \nTypeMadeStyleCoinve{}).

\begin{table}[t]
\centering
\small
\caption{Editing accuracy of redo turns, by whether the undo before them held. Held: the share of redos
whose undo held. The other columns give the accuracy on all redos, where the undo held, and where it failed
(turns in parentheses). The baselines pass a redo mostly when the undo before it had failed.}
\label{tab:redo-held}
\setlength{\tabcolsep}{0pt}
\begin{tabular*}{\columnwidth}{@{\extracolsep{\fill}}lcccc@{}}
\toprule
& Held & All & Undo held & Undo failed \\
\midrule
\input{generated/tab_mt_redo_held.tex} 
\bottomrule
\end{tabular*}
\end{table}

\noindent\textbf{Redo after a failed undo.}
A redo is scored on the redone edit's questions, so a system whose undo failed passes the redo without doing
anything. To check whether a redo score reflects a real redo, Table~\ref{tab:redo-held} splits the
\nBhRedo{} redos by whether their undo held. An undo held at the turn before if
every true/false question on the undone edit (\nRedoUndoQMin{} or \nRedoUndoQMax{} per edit, from its
reverted, absent, and replaced sets) passes. A lenient rule (one passing question suffices) gives
the same picture: the baselines redo at \nRedoIFHeldLenientBaseMin{} to \nRedoIFHeldLenientBaseMax{} where the
undo held.

\subsection{Results by Turn}
\label{app:mt-vq}

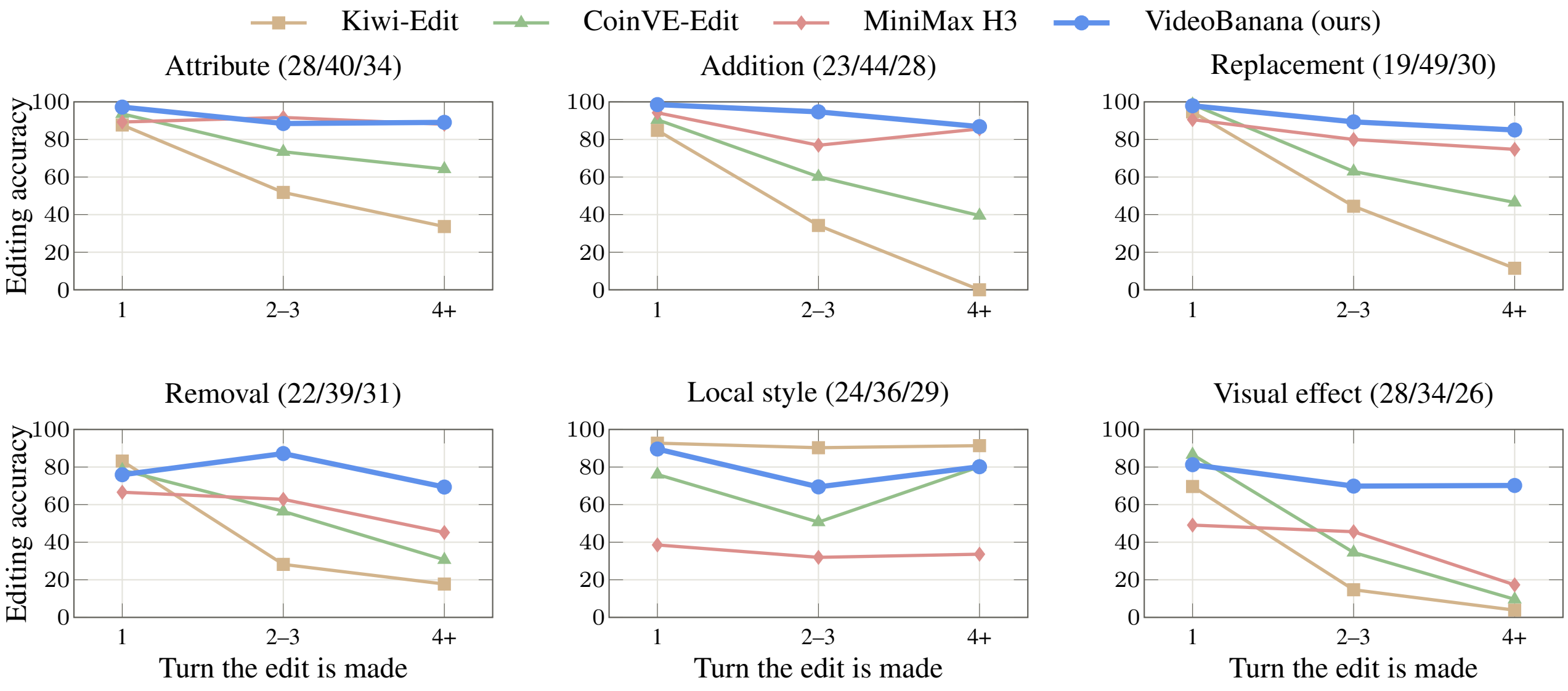
\begin{figure*}[t]
\centering
\begin{tikzpicture}
\begin{groupplot}[group style={group size=3 by 2, horizontal sep=1.25cm, vertical sep=1.5cm}, mtpanel,
  height=3.6cm, xmin=0.7, xmax=3.3, ymin=0, ymax=100, xtick={1,2,3}, xticklabels={1,2--3,4+}]
\nextgroupplot[title={Attribute (\nTypeTurnEditsAttr)}, ylabel={Editing accuracy}, legend to name=typelegend, legend columns=4, legend style={font=\small, draw=none, column sep=8pt}]
\addplot[mtkiwi] table[x=x, y=Kiwi] {generated/fig_mt_type_attr.dat}; \addlegendentry{Kiwi-Edit}
\addplot[mtcoinve] table[x=x, y=Coinve] {generated/fig_mt_type_attr.dat}; \addlegendentry{CoinVE-Edit}
\addplot[mththree] table[x=x, y=Hthree] {generated/fig_mt_type_attr.dat}; \addlegendentry{MiniMax H3}
\addplot[mtagent] table[x=x, y=Agent] {generated/fig_mt_type_attr.dat}; \addlegendentry{\agentname{} (ours)}
\nextgroupplot[title={Addition (\nTypeTurnEditsAdd)}]
\addplot[mtkiwi] table[x=x, y=Kiwi] {generated/fig_mt_type_add.dat};
\addplot[mtcoinve] table[x=x, y=Coinve] {generated/fig_mt_type_add.dat};
\addplot[mththree] table[x=x, y=Hthree] {generated/fig_mt_type_add.dat};
\addplot[mtagent] table[x=x, y=Agent] {generated/fig_mt_type_add.dat};
\nextgroupplot[title={Replacement (\nTypeTurnEditsRepl)}]
\addplot[mtkiwi] table[x=x, y=Kiwi] {generated/fig_mt_type_repl.dat};
\addplot[mtcoinve] table[x=x, y=Coinve] {generated/fig_mt_type_repl.dat};
\addplot[mththree] table[x=x, y=Hthree] {generated/fig_mt_type_repl.dat};
\addplot[mtagent] table[x=x, y=Agent] {generated/fig_mt_type_repl.dat};
\nextgroupplot[title={Removal (\nTypeTurnEditsRemove)}, ylabel={Editing accuracy}, xlabel={Turn the edit is made}]
\addplot[mtkiwi] table[x=x, y=Kiwi] {generated/fig_mt_type_remove.dat};
\addplot[mtcoinve] table[x=x, y=Coinve] {generated/fig_mt_type_remove.dat};
\addplot[mththree] table[x=x, y=Hthree] {generated/fig_mt_type_remove.dat};
\addplot[mtagent] table[x=x, y=Agent] {generated/fig_mt_type_remove.dat};
\nextgroupplot[title={Local style (\nTypeTurnEditsStyle)}, xlabel={Turn the edit is made}]
\addplot[mtkiwi] table[x=x, y=Kiwi] {generated/fig_mt_type_style.dat};
\addplot[mtcoinve] table[x=x, y=Coinve] {generated/fig_mt_type_style.dat};
\addplot[mththree] table[x=x, y=Hthree] {generated/fig_mt_type_style.dat};
\addplot[mtagent] table[x=x, y=Agent] {generated/fig_mt_type_style.dat};
\nextgroupplot[title={Visual effect (\nTypeTurnEditsEffect)}, xlabel={Turn the edit is made}]
\addplot[mtkiwi] table[x=x, y=Kiwi] {generated/fig_mt_type_effect.dat};
\addplot[mtcoinve] table[x=x, y=Coinve] {generated/fig_mt_type_effect.dat};
\addplot[mththree] table[x=x, y=Hthree] {generated/fig_mt_type_effect.dat};
\addplot[mtagent] table[x=x, y=Agent] {generated/fig_mt_type_effect.dat};
\end{groupplot}
\node[anchor=south] at ($(group c2r1.north)+(0,0.45cm)$) {\pgfplotslegendfromname{typelegend}};
\end{tikzpicture}
\caption{Editing accuracy of follow-ups by edit type and by the turn that makes them (turn 1,
turns 2--3, turn 4 or later), each scored at that turn on the edit's own questions, as in
Table~\ref{tab:mt-ops}. In parentheses: number of edits in the three groups.}
\label{fig:mt-types-by-turn}
\end{figure*}

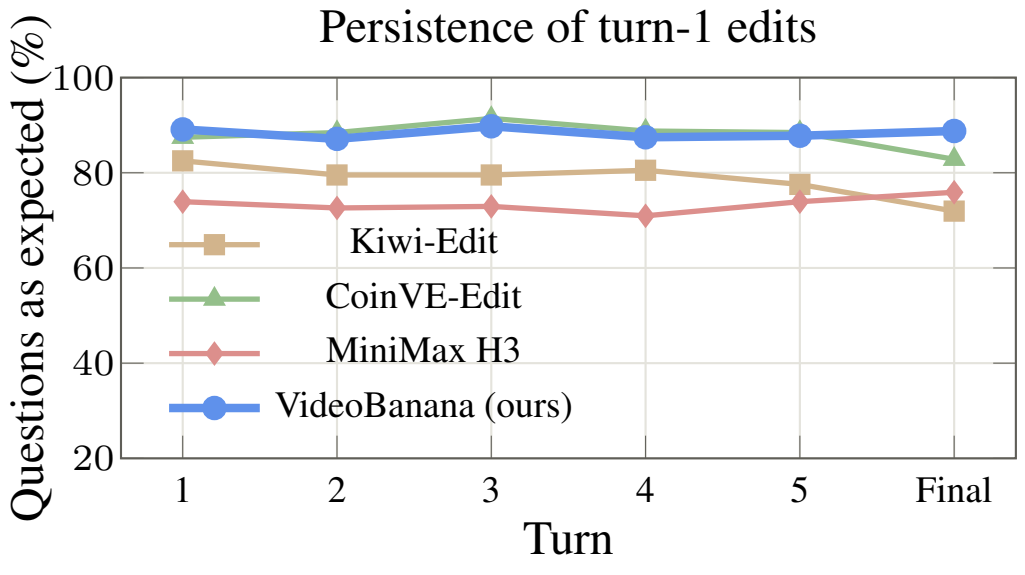
\begin{figure}[t]
\centering
\begin{tikzpicture}
\begin{axis}[mtpanel, width=0.9\columnwidth, title={Persistence of turn-1 edits}, xlabel={Turn},
  ylabel={Questions as expected (\%)}, xmin=0.6, xmax=6.4, ymin=20, ymax=100, xtick={1,...,6},
  xticklabels={1,2,3,4,5,Final}, legend pos=south west, legend style={font=\scriptsize, draw=none, fill=none}]
\addplot[mtkiwi] table[x=x, y=Kiwi] {generated/fig_mt_survival.dat}; \addlegendentry{Kiwi-Edit}
\addplot[mtcoinve] table[x=x, y=Coinve] {generated/fig_mt_survival.dat}; \addlegendentry{CoinVE-Edit}
\addplot[mththree] table[x=x, y=Hthree] {generated/fig_mt_survival.dat}; \addlegendentry{MiniMax H3}
\addplot[mtagent] table[x=x, y=Agent] {generated/fig_mt_survival.dat}; \addlegendentry{\agentname{} (ours)}
\end{axis}
\end{tikzpicture}
\caption{The \nSurvEdits{} edits made at turn 1 that stay requested until the end of their session.}
\label{fig:mt-survival}
\end{figure}

\begin{table}[t]
\centering
\small
\caption{Scores of each system at every turn, as plotted in Fig.~\ref{fig:mt-turns}, with TQ added. \agentname{} has the highest editing accuracy and content preservation at every turn.}
\label{tab:mt-per-turn}
\setlength{\tabcolsep}{3pt}
\resizebox{\columnwidth}{!}{%
\begin{tabular}{@{}lrrrrrrrr@{}}
\toprule
Turn & 1 & 2 & 3 & 4 & 5 & 6 & 7 & 8 \\
\midrule
\input{generated/tab_mt_turns_all.tex} 
\bottomrule
\end{tabular}}
\end{table}

Table~\ref{tab:mt-per-turn} gives the scores of each system at every turn. Editing accuracy and content
preservation are over all questions of the turn. Turns 6--8 exist in \nSessionsTurnSix{},
\nSessionsTurnSeven{}, and \nSessionsTurnEight{} sessions.

\noindent\textbf{Technical quality across turns.}
Every editor raises TQ above the source with its first edit (by \nTurnOneDTQHthree\% to
\nTurnOneDTQCoinve\%). TQ is DOVER++'s technical quality~\cite{dover}, which penalizes the motion blur and
noise of the real VEFX-Bench footage that regenerated frames lack. \agentname{} stays above
the source at every turn. Chained, CoinVE-Edit and MiniMax H3 fall below it from turn
\nTQBelowFromCoinve{} and Kiwi-Edit from turn \nTQBelowFromKiwi{}.

\noindent\textbf{Physical commonsense across turns.}
PC is lower at the final turn than at turn 1 for every system, \agentname{} included (\nTurnOnePCAgent{} to
\nFinalAbsPCAgent{}), and \agentname{} stays the highest. The drop comes with the edits themselves: editing
all requests once from the source already reaches \nFinalPCFromSource{}, against \nSourcePC{} for the
unedited source clips.

\noindent\textbf{Persistence of earlier edits.}
The baselines fall behind mainly on new requests. They keep most of what they made early: CoinVE-Edit still
answers \nSurvCoinveFinal\% of the questions on its turn-1 edits as expected at the final turn, and only
Kiwi-Edit loses a clear part of them (\nSurvKiwiOne\% to \nSurvKiwiFinal\%; Fig.~\ref{fig:mt-survival}). For
edits made at turns 2--5, no group loses 6 points or more by the final turn. What falls is the score
of a new edit.

\noindent\textbf{Decline by edit type.}
Kiwi-Edit and CoinVE-Edit make every type except local style at least 20 points worse at turn 4 or later
than at turn 1 (Fig.~\ref{fig:mt-types-by-turn}). \agentname{} loses less than 15 points on every edit type.

\subsection{Controlled Comparisons}
\label{app:controlled}

\begin{table}[t]
\centering
\small
\caption{Number of active turns on which each part of Table~\ref{tab:mt-ablation} raises / keeps / lowers
the score. Each part raises both scores on more turns than it lowers them.}
\label{tab:mt-controlled}
\setlength{\tabcolsep}{0pt}
\begin{tabular*}{\columnwidth}{@{\extracolsep{\fill}}lrcc@{}}
\toprule
& Turns & EA & CP \\
\midrule
\input{generated/tab_mt_controlled.tex} 
\bottomrule
\end{tabular*}
\end{table}

\begin{table}[t]
\centering
\small
\caption{Editing the previous output (Prev) or an earlier checkpoint (Tree), on the turns where Tree goes back. Cells are Prev / Tree. Reused: the checkpoint is the output. Edited: the editor is called on it again. Tree is higher overall in SA, EA, and CP.}
\label{tab:tree-verifier}
\setlength{\tabcolsep}{0pt}
\begin{tabular*}{\columnwidth}{@{\extracolsep{\fill}}lrccc@{}}
\toprule
& Turns & SA & EA & CP \\
\midrule
Reused & \nOneStepReuseTurns & \nOneStepReuseSADev{} / \textbf{\nOneStepReuseSAFirst} & \nOneStepReuseIFDev{} / \textbf{\nOneStepReuseIFFirst} & \nOneStepReuseCPDev{} / \textbf{\nOneStepReuseCPFirst} \\
Edited & \nOneStepBackTurns & \nOneStepBackSADev{} / \textbf{\nOneStepBackSAFirst} & \textbf{\nOneStepBackIFDev}{} / \nOneStepBackIFFirst & \nOneStepBackCPDev{} / \textbf{\nOneStepBackCPFirst} \\
\midrule
All & \nOneStepTurns & \nOneStepSADev{} / \textbf{\nOneStepSAFirst} & \nOneStepIFDev{} / \textbf{\nOneStepIFFirst} & \nOneStepCPDev{} / \textbf{\nOneStepCPFirst} \\
\bottomrule
\end{tabular*}
\end{table}

\noindent\textbf{Dependence between the parts.}
The three parts depend on one another. Goal tracking depends on entry selection: an undone
edit can be left out only if the turn starts from a video without it. If every turn edits the previous
output, the goal cannot change what that video already shows. Entry selection depends on the verifier and
on goal tracking: it reads the verifier's scores to know what each stored output shows, and it compares
them with the goal to decide what an entry preserves and what is left to execute. The verifier depends on
goal tracking: it scores every intent of the goal, and without a goal it has nothing to check.

Removing one part alone therefore leaves another part without its input. Without the verifier, no stored
output has scores, and entry selection returns the source at every turn (Sec.~\ref{sec:tree}). Without goal
tracking, there is no goal to select an entry for or to verify against, and every message has to be
executed. Without entry selection, every turn edits the previous output, and the goal cannot leave out an
undone edit. Each case switches off more than one part. Table~\ref{tab:mt-ablation} therefore compares each
part with everything else held fixed, on the turns where the part is active. Table~\ref{tab:mt-controlled}
counts how often the part helps on these turns. Turns without a preservation question do not count for content preservation.

\noindent\textbf{Comparison for entry selection.}
The active turns $t$ are those where the goal sends \agentname{} back to an earlier checkpoint, $V_0$ or
$Y_j$ with $j<t-1$. Both sides start from \agentname{}'s own run up to the previous turn. With entry
selection off, the turn edits the previous output $Y_{t-1}$ (Prev). With it on, it is \agentname{}'s first
sample from the selected entry (Tree). Both use the same goal, the same instruction writer, one sample, and
no verifier. Table~\ref{tab:tree-verifier} splits these turns. Prev loses most on the reused turns: most of
them take out an earlier edit, and the editor often leaves part of it. On the turns that edit again from an
earlier checkpoint, both reach about the same EA, and Tree preserves more content.

\noindent\textbf{Comparison for goal tracking.}
The active turns are the final turns of the \nCtlGoalActiveSessions{} sessions in which an edit was
undone or replaced. Both sides run \agentname{} with $D=1$ and no verifier, so each makes one editor call
on the source. With goal tracking off, the call gets every user message. With it on, it gets the
instructions written from the tracked goal.

\noindent\textbf{Comparison for the verifier.}
The active turns are the \nCtlVerifActiveTurns{} turns where \agentname{} commits its second sample and its
first attempt gave a video. Both sides are \agentname{}'s run with the same entry. With the verifier
off, the first sample is the output. Elsewhere, \agentname{} commits its first sample or reuses a
stored output, so the sides match.

\noindent\textbf{Turns covered by each comparison.}
The comparisons do not all cover the same turns. Entry selection and the verifier are decisions of a single
turn. Each can be switched off at one turn while the earlier turns stay as they are, so both are measured
on all their active turns. Goal tracking summarizes the whole conversation. The side without it is run at
the final turn, where the conversation is complete.

\subsection{Analysis of Verifier Rejections}
\label{app:gate-reasons}

\begin{table}[t]
\centering
\small
\caption{Conditions that a first sample fails when the verifier rejects it, over the \nGateRejected{}
rejected first samples. A sample can fail several conditions.}
\label{tab:gate-reasons}
\setlength{\tabcolsep}{0pt}
\begin{tabular*}{\columnwidth}{@{\extracolsep{\fill}}lrr@{}}
\toprule
Failed condition & Samples & Share \\
\midrule
\input{generated/tab_gate_reasons.tex} 
\bottomrule
\end{tabular*}
\end{table}

The verifier checks the edits, the rest of the video, and the visual quality (Sec.~\ref{sec:verify}). In
Table~\ref{tab:mt-ablation} it still raises editing accuracy much more than content preservation.
\agentname{} judged \nGateJudged{} first samples and rejected \nGateRejected{} of them
(Table~\ref{tab:gate-reasons}). Three things explain the difference.
\begin{tightitemize}
    \item \emph{Most conditions concern edits.} The gate has one condition for each edit of the turn and one
          for each earlier edit to keep, on average \nGateEditsPerTurn{} and \nGateKeptPerTurn{} per turn. It
          has one condition for the rest of the video and one for the visual quality.
    \item \emph{Edits fail more often.} An edit of the turn is missing in $\nGateEditFailPct\%$ of the first
          samples. The rest of the video fails in $\nGateRestFailPct\%$. The gate tolerates small changes
          there: \nGateRestOneBelow{} samples are scored one level below the best on the rest of the video
          and pass. Of the rejections, $\nGateOnlyEditPct\%$ are for a missing edit alone and
          $\nGateOnlyRestPct\%$ for the rest of the video alone.
    \item \emph{Part of the gain counts as editing accuracy.} Scope accuracy asks whether an edit is clean
          and confined to its target. An edit that spills over its target lowers scope accuracy, not content
          preservation.
\end{tightitemize}

\subsection{Sensitivity to the Depth Limit}
\label{app:depth-limit}

\begin{table}[t]
\centering
\small
\caption{Mean final-turn scores under each depth limit $D$, and a per-session comparison that counts the
sessions in which each setting scores higher. $D=3$ has the higher mean EA, and ties are most common per
session, except for CP.}
\label{tab:depth-limit}
\setlength{\tabcolsep}{0pt}
\begin{tabular*}{\columnwidth}{@{\extracolsep{\fill}}lccccc@{}}
\toprule
& SA & SPA & EP & EA & CP \\
\midrule
\input{generated/tab_depth_limit.tex} 
\bottomrule
\end{tabular*}
\end{table}

\begin{table}[t]
\centering
\small
\caption{Mean SSIM to the previous output under each depth limit $D$, on the turns where $D=3$ starts
from $Y_{t-1}$ or an earlier output $Y_j$. The SSIM is higher with $D=3$ for both entries.}
\label{tab:depth-ssim}
\setlength{\tabcolsep}{0pt}
\begin{tabular*}{\columnwidth}{@{\extracolsep{\fill}}lcccccc@{}}
\toprule
& \multicolumn{3}{c}{Frame} & \multicolumn{3}{c}{Edits} \\
\cmidrule(lr){2-4}\cmidrule(lr){5-7}
& $Y_{t-1}$ & $Y_j$ & All & $Y_{t-1}$ & $Y_j$ & All \\
\midrule
\input{generated/tab_depth_ssim.tex} 
\bottomrule
\end{tabular*}
\end{table}

$D$ bounds the generation depth of every output (Sec.~\ref{sec:tree}). It is a setting of entry selection.
To see whether the results of \agentname{} hold at the edge of its range, where entry selection has the
least freedom, we run \agentname{} with $D=1$, the smallest value, on all \nSessions{} sessions and compare
it with the default $D=3$ on the same sessions.

\noindent\textbf{Compared settings.}
With $D=1$, a turn returns a stored output that already shows the goal, or it executes the whole goal on the
source. Everything else is unchanged: the goal is tracked, every output is verified and stored with its
scores, and a stored output is reused when the goal returns to an earlier state. Only continuing from a
stored output is switched off. With $D=3$, a turn that edits starts from the previous output, an earlier
output, or the source (Sec.~\ref{sec:tree}). The settings thus differ where $D=3$
starts from a stored output. They are \nDepthStoredTurns{} of the \nDepthLaterTurns{} turns from turn 2 on:
\nDepthEntryPrevTurns{} start from the previous output and \nDepthEntryEarlierTurns{} from an earlier one.

\noindent\textbf{Effect on the final-turn scores.}
With $D=3$, the mean editing accuracy is slightly higher, \nDepthThreeIF{} against \nDepthOneIF{} with
$D=1$ (Table~\ref{tab:depth-limit}). The gain is in scope accuracy and edit persistence. Semantic accuracy
and content preservation are within one point for the two settings. Per session, a tie is the most common
outcome: editing accuracy is the same in \nDepthCountIFSame{} of the \nDepthSessions{} sessions, higher
with $D=3$ in \nDepthCountIFThreeHigher{}, and higher with $D=1$ in \nDepthCountIFOneHigher{}. Content
preservation is higher with $D=1$ in \nDepthCountCPOneHigher{} sessions and with $D=3$ in
\nDepthCountCPThreeHigher{}. Scope accuracy is compared on the \nDepthCountSPAN{} sessions and content
preservation on the \nDepthCountCPN{} sessions whose final turn has such a question.

\noindent\textbf{Effect on consistency across turns.}
The larger difference is in consistency across turns (Table~\ref{tab:depth-ssim}). On
\nDepthCountFrameThreeHigher{} of the \nDepthStoredTurns{} turns where the two settings differ, the output with $D=3$
is closer to the previous output, the one the user has just seen. Inside edits the previous output
already showed, this holds on \nDepthCountEditsThreeHigher{} of \nDepthRegionTurns{} turns. The mean SSIM
is higher with $D=3$ for both kinds of stored entry. The checklist does not see this difference, because it
compares each output with the goal. It asks whether the jacket is blue, and none asks whether it is
the jacket of the previous turn. With $D=1$, an edit kept from earlier turns is regenerated at every
turn. It can pass the checklist every turn and still look different each time.

\noindent\textbf{Measurement of consistency.}
SSIM compares the output of a turn with the output of the turn before, on 8 uniformly sampled frames in
grayscale, resized to a width of 640 pixels, with an $11\times11$ Gaussian window ($\sigma=1.5$). Frame is
the mean over the whole frame. Edits is the mean inside the attribute, replacement, and removal edits that
both outputs should show, excluding the turn's new edit. The region of an edit is CoinVE-Edit's
localization of its instruction on the source. \nDepthRegionTurns{} of the \nDepthStoredTurns{} turns
keep such an edit: \nDepthEntryPrevEditTurns{} start from the previous output and
\nDepthEntryEarlierEditTurns{} from an earlier one.

\noindent\textbf{Choice of $D$.}
With $D=1$, every edited turn is a new sample from the source. The edits kept from earlier turns are
executed again, and the rest of the video is generated again. Both can then differ from the previous
output. A larger $D$ lets a turn execute only what is new, on an output the user has already seen. The
outputs of consecutive turns are then more similar (Table~\ref{tab:depth-ssim}). Editing the previous
output at every turn is the other extreme. Consecutive outputs are then closest, but video quality falls
with every generation (Sec.~\ref{sec:exp-quality}), and an edit that the user has undone is often left in
part (Sec.~\ref{sec:exp-ablation}). $D$ bounds the depth between the two.
$D=3$ was set on the development conversations, before any benchmark run (App.~\ref{app:dev-data}). There,
CoinVE-Edit used turn by turn loses little video quality in its first two generations and more from the
third. $D$ was not chosen by accuracy on the benchmark. These conversations hardly overlap with the
benchmark: on the \nHeldSessions{} sessions whose clips were not used in development, the final-turn editing
accuracy of \agentname{} is \nHeldFinalIFAgent{}, against \nFinalIFAgent{} on all sessions.

\begin{figure*}[p]
\centering
\includegraphics[width=\textwidth]{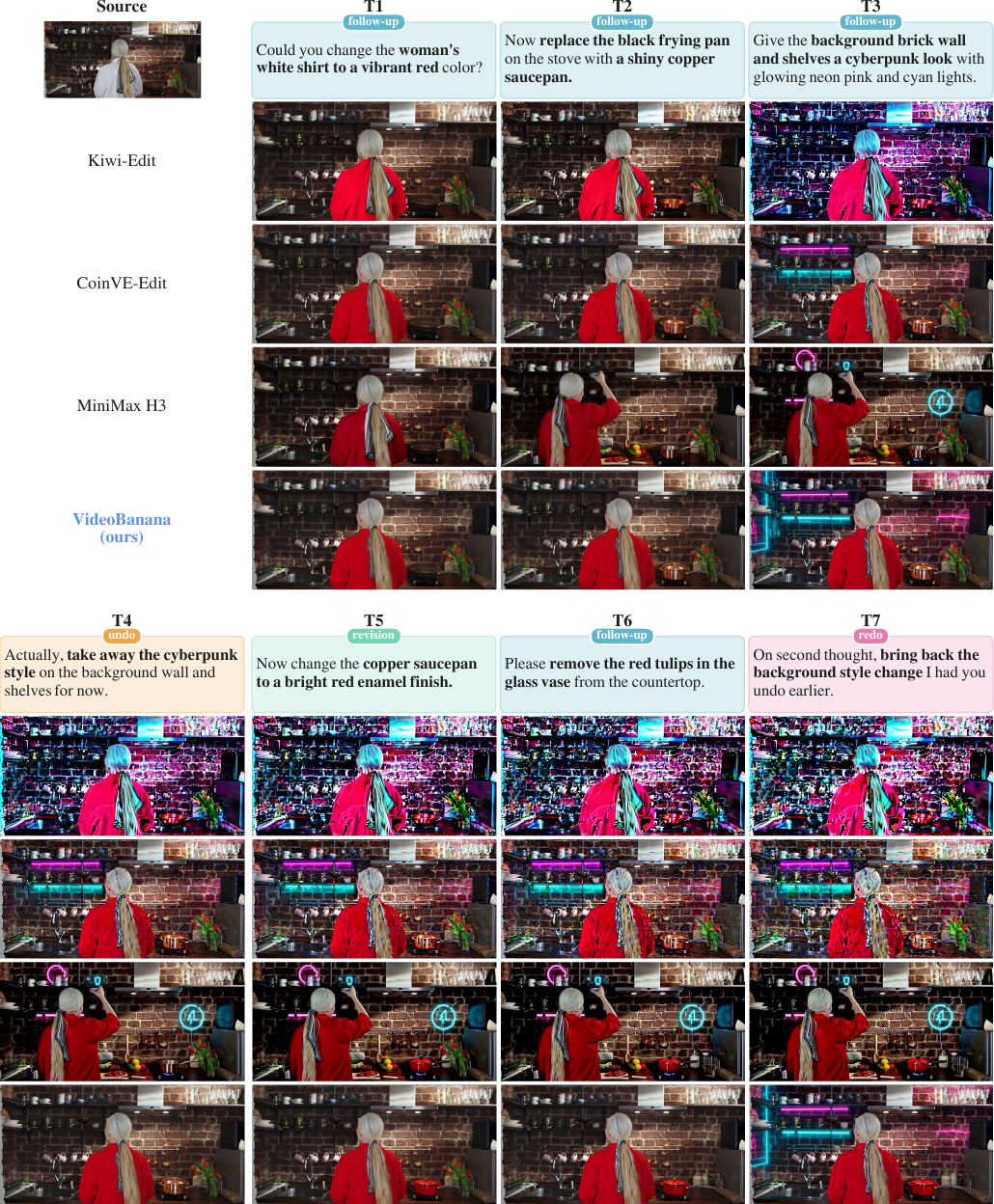}
\caption{Qualitative comparison on the session of Figure~\ref{fig:mt-architecture}. None of the baselines undoes
the cyberpunk style at T4. Kiwi-Edit and CoinVE-Edit also accumulate artifacts over the turns, and MiniMax H3
changes the viewpoint and the woman's pose from T2 on. \agentname{} undoes the style at T4, brings it back at
T7, and keeps the rest of the scene unchanged.}
\label{fig:mt-example-0076}
\end{figure*}

\begin{figure*}[p]
\begin{minipage}[t][0.97\textheight][c]{\textwidth}
\centering
\includegraphics[width=\textwidth]{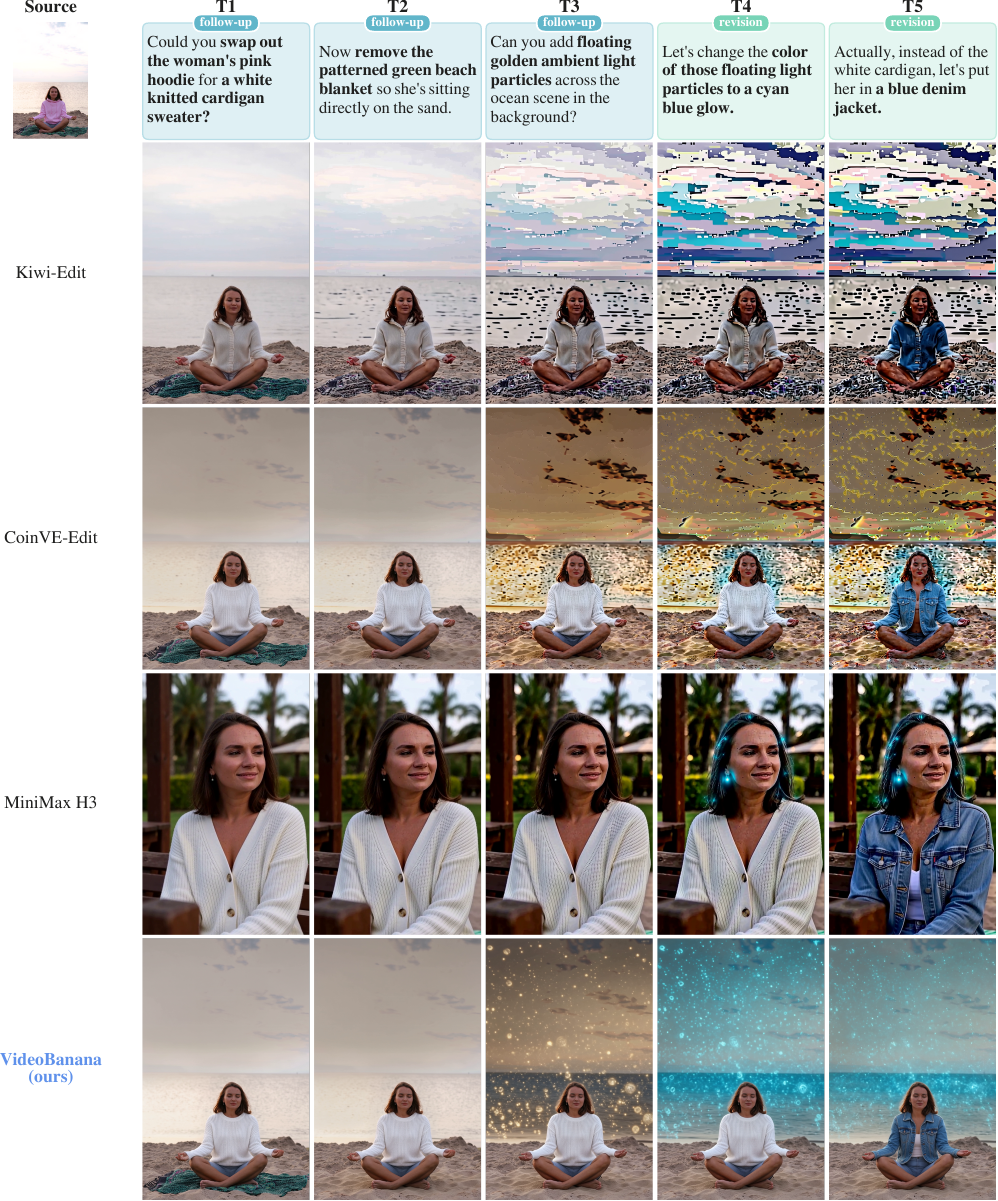}
\caption{Qualitative comparison on a simple session. MiniMax H3 changes the viewpoint. Kiwi-Edit and
CoinVE-Edit edit their own previous output at every turn, so artifacts accumulate. \agentname{} carries out
every request and keeps the rest of the scene.}
\label{fig:mt-example-0155}
\end{minipage}
\end{figure*}

\begin{figure*}[p]
\centering
\includegraphics[width=\textwidth]{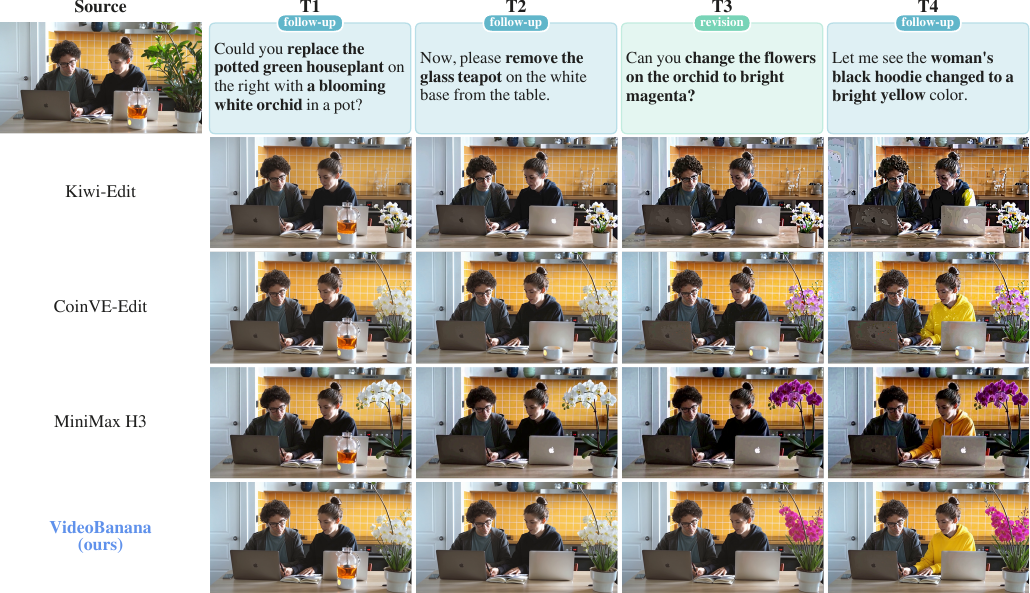}
\caption{T1--T4 of the session in Figure~\ref{fig:teaser}, expanded.}
\label{fig:mt-chain-full}
\vspace{6mm}
\includegraphics[width=\textwidth]{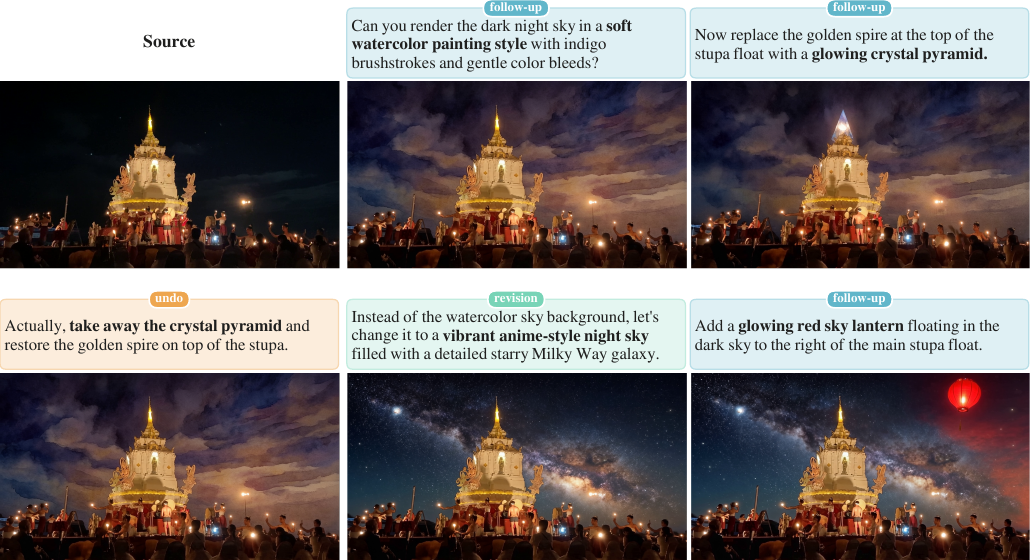}
\caption{More results of \agentname{}.}
\label{fig:mt-example-agent}
\end{figure*}

\begin{figure*}[p]
\centering
\includegraphics{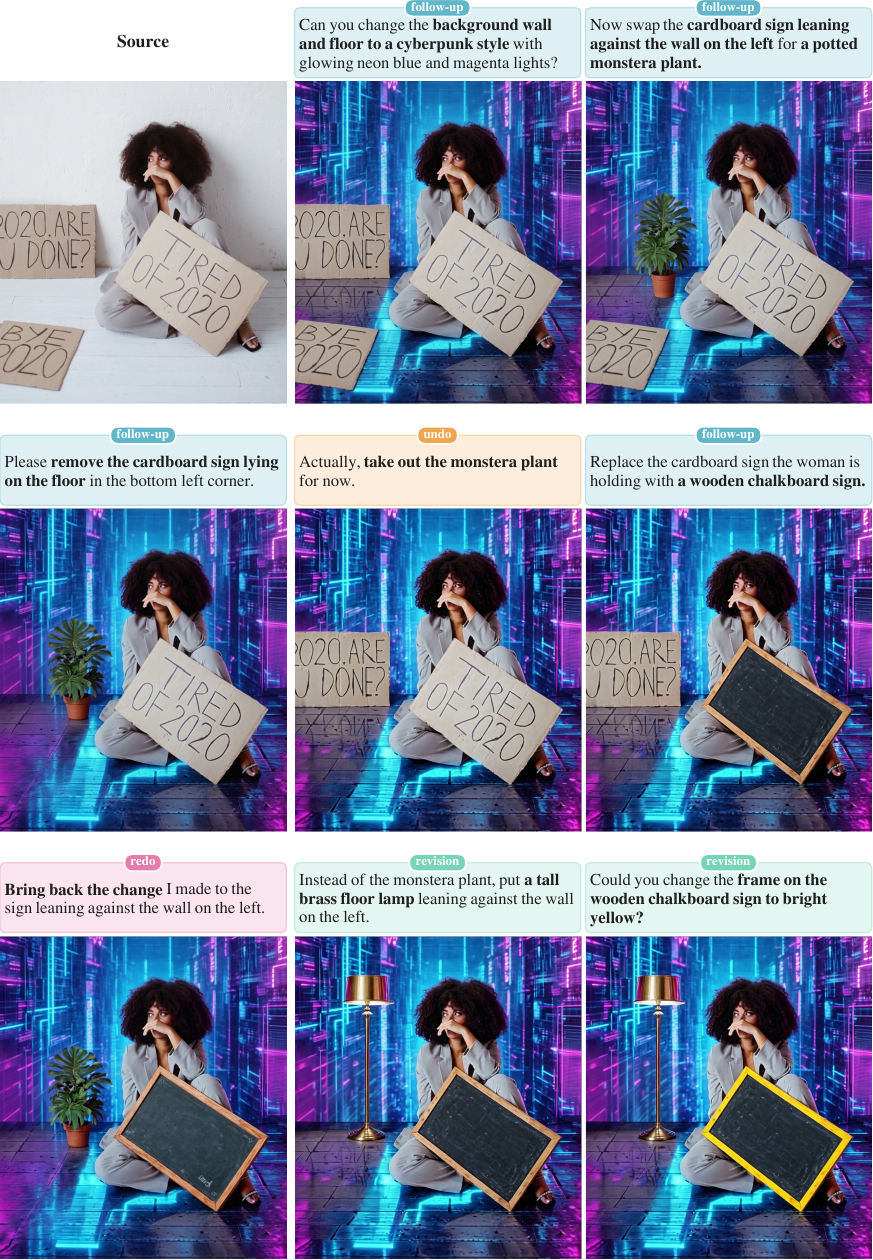}
\caption{More results of \agentname{}.}
\label{fig:mt-example-agent-2}
\end{figure*}

\begin{figure*}[t]
\centering
\includegraphics[width=\textwidth]{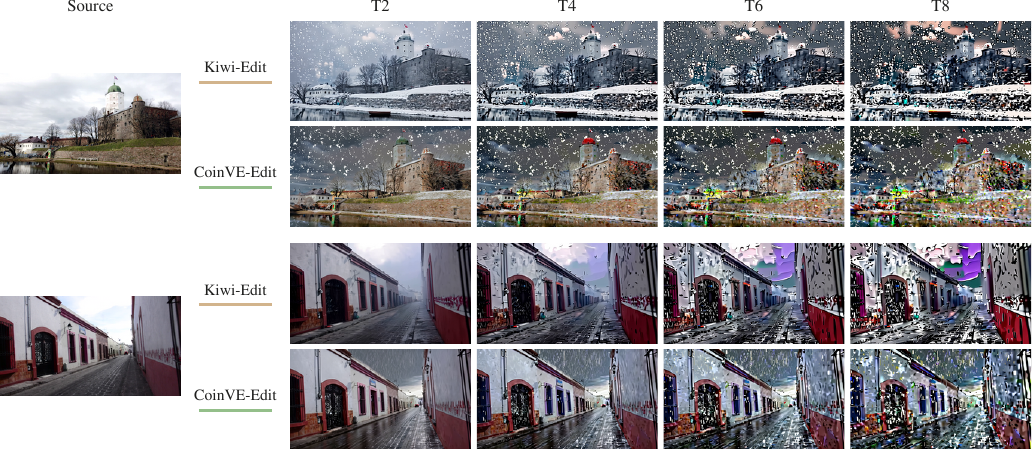}
\caption{\textbf{Loss of visual quality of Kiwi-Edit and CoinVE-Edit used turn by turn.} Two sessions of
eight turns. Every turn edits the previous output, so the artifacts of each generation build up over the turns.}
\label{fig:baseline-quality}

\vspace{5mm}
\small
\raggedright (a) MiniMax H3, session 0073, turn 1: ``Could you replace the silver thermos flask on the table with a red ceramic teapot?''\\[0.5mm]
\includegraphics[width=\textwidth]{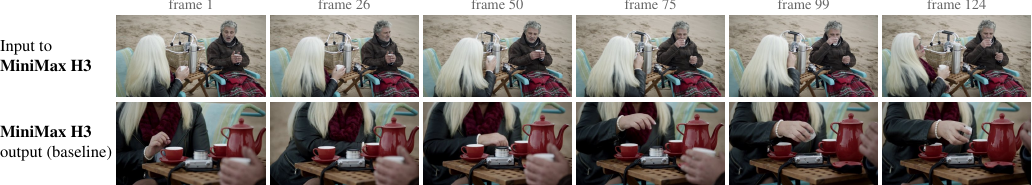}\\[2mm]
(b) MiniMax H3, session 0069, turn 1: ``Could you replace the open pizza box sitting on the sofa with a cute plush brown teddy bear?''\\[0.5mm]
\includegraphics[width=\textwidth]{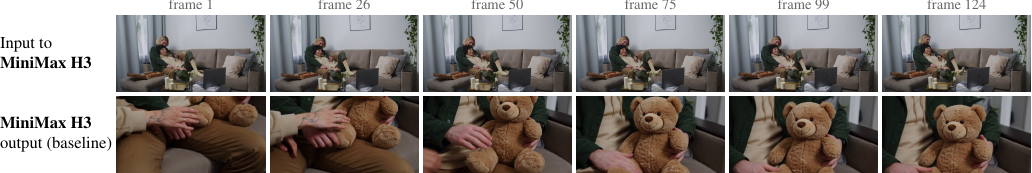}
\caption{\textbf{Change of viewpoint by MiniMax H3.} Turn 1 of two sessions.
In each pair, top: the 124-frame input; bottom: the output (same six frames). The output is a close-up
of the edited object and matches no part of the input.}
\label{fig:h3-view}
\end{figure*}

\begin{figure*}[p]
\centering
\includegraphics[width=\textwidth]{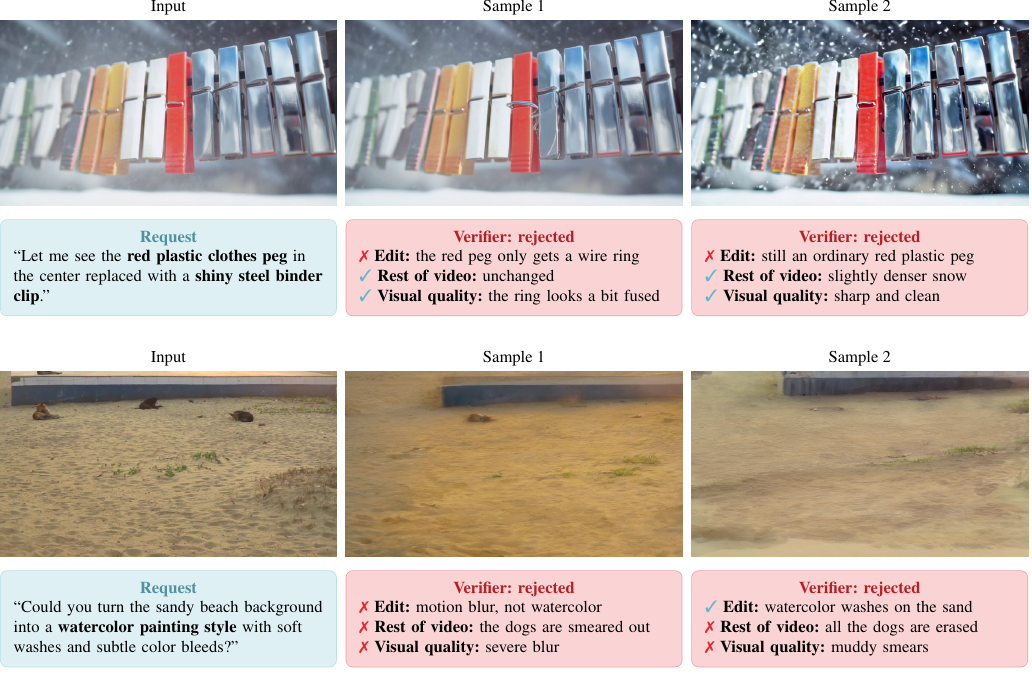}
\caption{\textbf{Failure cases of \agentname{} caused by its video editor.} CoinVE-Edit fails in every
sample, so \agentname{} fails as well. Top: the peg is left unchanged. Bottom: the style erases the dogs.}
\label{fig:agent-failure}
\end{figure*}

\section{Qualitative Results}
\label{app:qualitative}

\subsection{Example Sessions}
\label{app:mt-examples}

Figures~\ref{fig:mt-example-0076}, \ref{fig:mt-example-0155}, and~\ref{fig:mt-chain-full} compare all systems
on multi-turn sessions. Figure~\ref{fig:mt-example-0076} shows the session of Figure~\ref{fig:mt-architecture}, and
Figure~\ref{fig:mt-example-0155} a simple session; its frames leave out the unused sand at the bottom of the
video. Figure~\ref{fig:mt-chain-full} expands T1--T4 of the session in Figure~\ref{fig:teaser}.
Figures~\ref{fig:mt-example-agent} and~\ref{fig:mt-example-agent-2} show \agentname{}'s output after every turn of further sessions.
\supvideos

\subsection{Failure Cases of the Baselines}
\label{app:baseline-failures}

\noindent\textbf{Loss of visual quality.}
To show what the falling scores of the dedicated editors look like, Fig.~\ref{fig:baseline-quality} gives
two sessions. This is their most common failure when used turn by turn.
\begin{tightitemize}
    \item \emph{What happens.} Every turn edits the previous output. Each generation adds speckles, broken
          textures, and patches of saturated color, and these build up over the turns.
    \item \emph{Effect on the scores.} VQR falls at every turn after the second
          (Fig.~\ref{fig:mt-turns}c, Table~\ref{tab:mt-per-turn}), and editing accuracy with it.
    \item \emph{The two sessions.} They have eight turns and are shown at turns 2, 4, 6, and 8. They are
          close to the median final-turn VQR of the eight-turn sessions, not the worst cases.
          \agentname{} uses the same editor as the second row and keeps every output
          within \nBhMaxDepth{} generations of the source.
\end{tightitemize}

\noindent\textbf{MiniMax H3 changes the view.}
\phantomsection\label{app:h3-view}
To explain part of MiniMax H3's lower content preservation, Fig.~\ref{fig:h3-view} shows a failure of its
outputs that \agentname{} does not show.
\begin{tightitemize}
    \item \emph{What happens.} When a local edit is asked for in a wide shot, MiniMax H3 can return a
          close-up of the edited object. The edit is visible, but the framing, the camera, and the rest
          of the scene are replaced.
    \item \emph{Effect on the scores.} The editing questions can still pass; the preservation questions
          compare with the source and fail. Part of MiniMax H3's lower content preservation comes from
          such close-up outputs.
    \item \emph{The two sessions.} MiniMax H3 scores 0 and 17.5 on content preservation and 100 and 75
          on editing accuracy. \agentname{} scores 100 on both and keeps the view.
\end{tightitemize}

\subsection{Failure Cases of \agentname{}}
\label{app:agent-failures}

\agentname{} is bounded by its editor. When the editor cannot make an edit, resampling does not help, and
the verifier rejects every sample (Figure~\ref{fig:agent-failure}).
\begin{tightitemize}
    \item \emph{Edit not made (top).} The request replaces the red peg in the middle with a steel binder clip.
          Both samples keep the red peg; sample 1 only adds a wire ring to it. The rest of the video and the
          visual quality pass, and the verifier rejects both samples on the edit alone.
    \item \emph{Edit that destroys the scene (bottom).} The first turn of a session asks for the sandy beach in
          watercolor style; the dogs on the sand stay. Sample 1 smears the sand like motion blur. Sample 2
          shows the style, but it spreads over the whole frame and erases the dogs. Both fail the checks on
          the rest of the video and on visual quality.
\end{tightitemize}

%% file: generated/tab_qwen_dev_held.tex
Development & 40 & 61.0 & 83.7 & 75.0 & 81.4 & 74.8 & $-2.3$ \\
Held-out & 104 & 64.9 & 86.0 & 76.6 & 85.3 & 76.0 & $-0.7$ \\
\midrule
All & 144 & 63.8 & 85.3 & 76.2 & 84.2 & 75.7 & $-1.1$ \\

%% file: generated/tab_other_editor.tex
Kiwi-Edit & Chain & 35.3 & 55.9 & 46.7 & 69.3 & 907 \\
 & \agentname{} & 71.8 & 65.7 & 75.7 & 75.6 & 1431 \\
\midrule
CoinVE-Edit & Chain & 50.6 & 60.3 & 61.7 & 77.2 & 907 \\
 & \agentname{} & 84.2 & 75.7 & 86.2 & 83.1 & 1254 \\

%% file: generated/tab_st_main.tex
\mutedcell{Seedance 2.0$^\dagger$} & \mutedcell{85.3} & \mutedcell{87.7} & \mutedcell{88.1} & \mutedcell{93.2} & \mutedcell{95.8} & \mutedcell{92.9} & \mutedcell{93.9} & \mutedcell{91.6} & \mutedcell{4.47} & \mutedcell{19.55} & \mutedcell{4.41} & \mutedcell{0.62} & \mutedcell{--} \\
\mutedcell{Kling O3$^\dagger$} & \mutedcell{86.9} & \mutedcell{80.9} & \mutedcell{89.1} & \mutedcell{92.6} & \mutedcell{90.3} & \mutedcell{93.9} & \mutedcell{84.5} & \mutedcell{87.5} & \mutedcell{4.49} & \mutedcell{18.36} & \mutedcell{4.37} & \mutedcell{0.61} & \mutedcell{--} \\
\mutedcell{VACE$^\dagger$} & \mutedcell{4.0} & \mutedcell{17.2} & \mutedcell{6.5} & \mutedcell{26.7} & \mutedcell{13.8} & \mutedcell{15.2} & \mutedcell{87.8} & \mutedcell{38.5} & \mutedcell{4.05} & \mutedcell{17.59} & \mutedcell{4.11} & \mutedcell{0.62} & \mutedcell{--} \\
\mutedcell{Ditto$^\dagger$} & \mutedcell{34.7} & \mutedcell{36.4} & \mutedcell{40.9} & \mutedcell{36.0} & \mutedcell{47.8} & \mutedcell{38.5} & \mutedcell{52.0} & \mutedcell{43.3} & \mutedcell{3.59} & \mutedcell{17.24} & \mutedcell{3.96} & \mutedcell{0.67} & \mutedcell{--} \\
\mutedcell{VINO$^\dagger$} & \mutedcell{83.6} & \mutedcell{66.8} & \mutedcell{89.1} & \mutedcell{78.1} & \mutedcell{82.3} & \mutedcell{85.9} & \mutedcell{61.7} & \mutedcell{74.5} & \mutedcell{4.06} & \mutedcell{17.28} & \mutedcell{4.08} & \mutedcell{0.68} & \mutedcell{--} \\
\mutedcell{OmniWeaving$^\dagger$} & \mutedcell{59.7} & \mutedcell{55.9} & \mutedcell{61.1} & \mutedcell{54.5} & \mutedcell{66.0} & \mutedcell{65.1} & \mutedcell{75.1} & \mutedcell{65.3} & \mutedcell{3.79} & \mutedcell{17.95} & \mutedcell{3.84} & \mutedcell{0.62} & \mutedcell{--} \\
\mutedcell{Kiwi-Edit$^\dagger$} & \mutedcell{76.5} & \mutedcell{69.9} & \mutedcell{80.3} & \mutedcell{78.4} & \mutedcell{78.5} & \mutedcell{80.8} & \mutedcell{70.3} & \mutedcell{75.0} & \mutedcell{4.14} & \mutedcell{19.33} & \mutedcell{4.30} & \mutedcell{0.68} & \mutedcell{--} \\
\mutedcell{SAMA$^\dagger$} & \mutedcell{75.6} & \mutedcell{73.4} & \mutedcell{79.6} & \mutedcell{83.4} & \mutedcell{83.9} & \mutedcell{88.1} & \mutedcell{90.1} & \mutedcell{83.8} & \mutedcell{3.61} & \mutedcell{18.08} & \mutedcell{4.19} & \mutedcell{0.72} & \mutedcell{--} \\
\mutedcell{CoinVE-Edit$^\dagger$} & \mutedcell{88.0} & \mutedcell{89.5} & \mutedcell{89.6} & \mutedcell{91.9} & \mutedcell{91.2} & \mutedcell{95.3} & \mutedcell{90.8} & \mutedcell{90.9} & \mutedcell{4.13} & \mutedcell{19.57} & \mutedcell{4.31} & \mutedcell{0.72} & \mutedcell{--} \\
\midrule
CoinVE-Edit (re-run) & 86.7 & 87.6 & 90.3 & 90.6 & 91.2 & 94.0 & 87.3 & 89.1 & 4.33 & 20.26 & 4.30 & 0.72 & 361 \\
\agentname{} (ours) & \textbf{88.3} & \textbf{89.1} & \textbf{92.7} & \textbf{92.3} & \textbf{94.1} & \textbf{95.7} & \textbf{89.2} & \textbf{91.1} & 4.33 & 20.21 & 4.31 & 0.72 & 464 \\

%% file: generated/tab_st_acted.tex
Acc. & 93.3 \stzero & 95.7 \stzero & 89.6 \stzero & 92.9 \stzero \\
Rej.\,$\rightarrow$\,Acc. & 74.6 \stup{15.2} & 82.3 \stup{15.8} & 85.4 \stup{8.6} & 80.8 \stup{13.2} \\
Rej.\,$\rightarrow$\,Rej. & 76.7 \stdn{0.5} & 78.7 \stup{3.1} & 78.1 \stup{5.3} & 77.8 \stup{2.7} \\

%% file: generated/tab_st_all.tex
All & 88.2 \stup{1.9} & 92.0 \stup{2.0} & 87.3 \stup{1.9} & 89.1 \stup{2.0} \\

%% file: generated/tab_mt_ops_types.tex
Follow-up (564) & 48.7 & 62.5 & 69.2 & \textbf{86.2} \\
\quad Attribute (102) & 55.5 & 75.9 & 89.8 & \textbf{91.1} \\
\quad Addition (95) & 38.6 & 62.3 & 83.9 & \textbf{93.6} \\
\quad Replacement (98) & 45.1 & 65.6 & 80.2 & \textbf{90.0} \\
\quad Removal (92) & 37.4 & 52.8 & 57.6 & \textbf{78.5} \\
\quad Local style (89) & \textbf{91.3} & 67.1 & 34.3 & 78.4 \\
\quad Visual effect (88) & 29.0 & 43.8 & 38.4 & \textbf{73.6} \\
\midrule
Revision (152) & 23.8 & 35.8 & 62.0 & \textbf{81.6} \\
\quad Attribute (80) & 20.0 & 41.1 & 70.2 & \textbf{83.4} \\
\quad Addition (16) & 13.8 & 12.7 & \textbf{88.1} & 79.0 \\
\quad Replacement (13) & 26.4 & 31.4 & 58.9 & \textbf{86.6} \\
\quad Removal & -- & -- & -- & -- \\
\quad Local style (23) & 58.7 & 59.8 & 35.9 & \textbf{71.7} \\
\quad Visual effect (20) & 10.8 & 7.5 & 17.5 & \textbf{78.3} \\
\midrule
Undo (114) & 40.8 & 47.8 & 49.8 & \textbf{92.7} \\
\quad last step (56) & 41.4 & 66.0 & 62.3 & \textbf{89.8} \\
\quad earlier step (58) & 42.9 & 20.3 & 32.9 & \textbf{98.4} \\
\quad Attribute (24) & 49.6 & 45.2 & 44.4 & \textbf{97.8} \\
\quad Addition (24) & 30.7 & 73.9 & 29.5 & \textbf{85.4} \\
\quad Replacement (16) & 47.5 & 82.8 & 83.8 & \textbf{98.0} \\
\quad Removal (19) & 36.8 & 31.6 & 52.6 & \textbf{89.5} \\
\quad Local style (19) & 35.0 & 40.0 & \textbf{73.8} & 72.5 \\
\quad Visual effect (12) & 37.5 & 31.3 & 56.3 & \textbf{100.0} \\
\midrule
Redo (77) & 37.3 & 43.3 & 49.5 & \textbf{81.6} \\
\quad Attribute (19) & 23.8 & 36.0 & 48.3 & \textbf{88.1} \\
\quad Addition (14) & 41.2 & 53.2 & 64.2 & \textbf{89.7} \\
\quad Replacement (7) & 43.7 & 29.0 & 34.0 & \textbf{71.0} \\
\quad Removal (13) & 34.6 & 42.3 & 46.2 & \textbf{80.8} \\
\quad Local style (13) & \textbf{100.0} & 90.4 & 67.3 & 78.8 \\
\quad Visual effect (11) & 4.0 & 27.6 & 33.1 & \textbf{65.8} \\
\midrule
Macro average & 37.7 & 47.4 & 57.6 & \textbf{85.5} \\

%% file: generated/tab_mt_redo_held.tex
Kiwi-Edit & 49\% & 37.3 & 10.9 & 66.6 (39) \\
CoinVE-Edit & 43\% & 43.3 & 11.0 & 69.1 (44) \\
MiniMax H3 & 45\% & 49.5 & 26.5 & 68.5 (42) \\
\agentname{} (ours) & 95\% & 81.6 & 80.8 & 95.8 (4) \\

%% file: generated/tab_mt_turns_all.tex
\multicolumn{9}{@{}l}{\emph{Editing accuracy}} \\
Kiwi-Edit & 86.0 & 66.4 & 53.3 & 46.1 & 39.9 & 37.9 & 36.1 & 26.7 \\
CoinVE-Edit & 88.7 & 74.9 & 69.3 & 62.1 & 56.6 & 53.7 & 54.5 & 46.1 \\
MiniMax H3 & 74.8 & 72.7 & 71.4 & 69.3 & 65.4 & 67.9 & 66.2 & 58.5 \\
\agentname{} (ours) & \textbf{91.5} & \textbf{88.4} & \textbf{89.0} & \textbf{86.1} & \textbf{84.7} & \textbf{87.1} & \textbf{85.2} & \textbf{83.3} \\
\midrule
\multicolumn{9}{@{}l}{\emph{Content preservation}} \\
Kiwi-Edit & 88.0 & 78.4 & 70.3 & 65.2 & 63.5 & 58.2 & 55.5 & 52.9 \\
CoinVE-Edit & 92.2 & 88.2 & 82.3 & 75.6 & 70.3 & 64.2 & 59.5 & 58.9 \\
MiniMax H3 & 79.8 & 75.2 & 68.5 & 66.4 & 62.7 & 59.3 & 57.7 & 54.2 \\
\agentname{} (ours) & \textbf{93.7} & \textbf{88.7} & \textbf{85.1} & \textbf{83.5} & \textbf{79.4} & \textbf{77.9} & \textbf{78.1} & \textbf{76.0} \\
\midrule
\multicolumn{9}{@{}l}{\emph{VQR} (source: 4.08)} \\
Kiwi-Edit & 4.00 & 3.68 & 2.68 & 1.68 & 1.05 & 0.70 & 0.51 & 0.43 \\
CoinVE-Edit & 4.03 & 3.92 & 3.47 & 2.67 & 1.79 & 1.15 & 0.68 & 0.39 \\
MiniMax H3 & 4.03 & 4.01 & 3.94 & 3.77 & 3.54 & 3.24 & 2.91 & 2.53 \\
\agentname{} (ours) & \textbf{4.06} & \textbf{4.09} & \textbf{4.03} & \textbf{4.06} & \textbf{4.00} & \textbf{4.04} & \textbf{4.09} & \textbf{4.15} \\
\midrule
\multicolumn{9}{@{}l}{\emph{TQ} (source: 14.54)} \\
Kiwi-Edit & 14.95 & 15.39 & 14.36 & 13.23 & 11.84 & 10.97 & 9.62 & 9.74 \\
CoinVE-Edit & \textbf{15.91} & \textbf{16.36} & 15.73 & 14.63 & 13.36 & 12.21 & 10.66 & 10.34 \\
MiniMax H3 & 14.70 & 14.63 & 14.44 & 13.88 & 13.19 & 12.53 & 11.64 & 11.42 \\
\agentname{} (ours) & 15.79 & 16.25 & \textbf{16.42} & \textbf{16.29} & \textbf{16.62} & \textbf{16.97} & \textbf{16.56} & \textbf{16.12} \\

%% file: generated/tab_mt_controlled.tex
Entry selection & 141 & 70 / 39 / 32 & 75 / 37 / 25 \\
Goal tracking & 72 & 36 / 27 / 9 & 28 / 25 / 18 \\
Verifier & 252 & 158 / 64 / 30 & 85 / 89 / 75 \\

%% file: generated/tab_gate_reasons.tex
An edit of this turn is missing & 267 & $78\%$ \\
The rest of the video is damaged & 89 & $26\%$ \\
Visual quality & 86 & $25\%$ \\
An earlier edit is damaged & 33 & $10\%$ \\

%% file: generated/tab_depth_limit.tex
\multicolumn{6}{@{}l}{\emph{Mean}} \\
$D=1$ & 84.2 & 85.2 & 82.5 & 84.0 & 77.0 \\
$D=3$ & 83.8 & 87.7 & 84.5 & 85.3 & 76.2 \\
\midrule
\multicolumn{6}{@{}l}{\emph{Per-session comparison}} \\
$D=1$ & 34 & 12 & 27 & 38 & 55 \\
Tie & 80 & 101 & 83 & 66 & 37 \\
$D=3$ & 30 & 18 & 34 & 40 & 50 \\

%% file: generated/tab_depth_ssim.tex
$D=1$ & .730 & .743 & .736 & .729 & .723 & .726 \\
$D=3$ & .858 & .821 & .840 & .887 & .812 & .848 \\

%% file: prompts_appendix.tex
\section{Prompt Templates}
\label{app:prompts}

All model calls request JSON and validate the response against a strict schema. A malformed or
inconsistent answer is returned to the model with the error, once (the intent compiler: up to twice).
Before that, the client tries each call up to three times if it fails or the reply is not JSON; the
Qwen client turns off thinking on its last try.
The JSON payload of each call is described after its system prompt. Some examples in the prompts name
objects that also appear in benchmark clips, such as a coffee grinder; none is an edit or a message of
the benchmark.

\subsection{Intent Refiner}
\begin{lstlisting}[style=prompt]
You keep track of what a user currently wants in a multi-turn video editing conversation.

The user edits one source video over several turns, and each turn produces a new video. You do not see the videos. After the latest turn, write down the edit intents the next video should satisfy. Each intent is later turned into an editing instruction and checked on its own by a video judge, and the system reuses an earlier video when the intents it needs are already there. The guidance below serves that purpose; where a case is not covered, do what serves it best.

What the user wants
- Requests stay in effect until the user revises or cancels them, even if an earlier video may not have shown them correctly.
- Do not add edits the user did not ask for.

Splitting into intents
- Aim for one intent per thing the user asked to change, so that each can be judged, kept or undone on its own. Changing an existing wall's color and its texture is two intents; "add a blue brick wall" or "replace the SUV with a white sedan" is one, because the appearance describes the new element. Several elements in one request usually mean several intents, and separate requests should not be merged just to have fewer.
- Write each intent as a self-contained description of the result, understandable without the conversation, for example "A table stands to the right of the window and to the left of the lamp."
- Carry over the new information the user gave about that edit: where it goes, how it relates to other things (for instance that the particles must not cover the flowers), its size, material or placement.
- Avoid wording that only makes sense relative to the conversation history, such as "undo", "instead" or "again".

Revisions and undo
- When the user changes something an active intent already sets, including the look of something added earlier, write a new intent with the updated description and list the old one in "replaces".
- Write a new intent only when the user changes that element itself. An intent that mentions another element (mist that must not cover the sedan) stays as it is when that other element changes or disappears; keep its id.
- When the user undoes a change, removes something that was added, or asks for the original back, make the corresponding intents inactive. Undoing a revision brings back the intent it replaced. There is no need for an intent that describes the original video.
- If the goal returns to a state an earlier intent already describes, reuse that intent's id instead of writing a new one: this is what lets the system go back to an earlier video. "Red" after "bright red" is usually the same state unless the user makes clear it is different.
- Intents that only make sense for an element that is gone (removed, replaced, or its addition undone) become inactive as well.

Constraints
- Separately list requests that add no new information and only stress that something should stay as it would anyway, such as "keep the lamp's light unchanged", "keep the pajamas' cut" or "leave the other cars alone". Anything that adds new information belongs in an intent instead.
- Keep this list current and drop items that no longer apply.
\end{lstlisting}

The payload lists every user turn so far with its index, every recorded intent with its id,
text, status (active or inactive) and the intents it replaced, and the output fields. The model
returns the changes of the latest turn (new intents with a local key, deactivated and reactivated
ids, each with a reason), the complete active list, and the preservation-only constraints;
\agentname{} checks that the changes lead from the previous active list to the stated one and assigns
permanent ids.

\subsection{Instruction Writer}
\begin{lstlisting}[style=prompt]
You write instructions for a video editing model.

The editor receives one starting video and a short list of instructions, and applies each instruction on its own, predicting a separate edit region for each. Your input is a list of target states the edited video must show, some notes on what the starting video already shows, and, for context, the user's original words. Turn the target states into instructions for this starting video.

- Write each instruction as one concrete imperative sentence describing the change to make to the starting video, for example "Change the black outer casing of the coffee grinder to dark blue." or "Add a bouquet of red roses on the counter between the amp on the left and the grinder."
- Phrase the change from what the starting video actually shows. If it already shows an earlier version of the same element (listed under "starting_video_shows"), change that version; if the target is an element of the original video, describe it as it appears there.
- Keep every detail of the target state: position, relation to other things (for example that particles must not cover the flowers), size, material, placement. The user's words can help recover a detail the target state left out, but do not add anything beyond the target state; the user's message may also contain requests that are handled elsewhere.
- The editor knows nothing about the conversation, so write only about the video: avoid references such as "again", "instead", "back" or "undo".
- Targets on the same element can share one instruction ("Replace the dark SUV with a red sedan."). Keep separate elements in separate instructions, and cover every target exactly once.
- Do not write instructions for what the starting video already shows correctly.
- Some inputs may list "edits_to_remove": edits the starting video shows that are no longer wanted. Write the change that removes each one, returning that part to how the original video looks unless a target says otherwise; such a removal can share an instruction with a target on the same element ("Change the dark blue casing of the coffee grinder to white.").
\end{lstlisting}

The payload states whether the entry is the source or an earlier output, what the entry already
shows (preserved intents and overwritten earlier versions), each target with its state and the
user words that introduced it, and the edits to remove (only when entry selection is off, App.~\ref{app:controlled}).
Every target must be covered by exactly one instruction.

\subsection{Verifier}

The system prompt is the single-turn verifier prompt (Appendix~\ref{app:prompt-verifier}) followed
by this addendum. Video 1 may already show earlier edits, so two lines of the single-turn prompt are
rewritten: Video 1 is ``The video the editor received for this step (it may already contain edits from
earlier steps)'', and score 1 of Instructional Following is ``Video 2 does not show the requested state at
all, or shows something unrelated to it. [\dots] Whether Video 2 differs from Video 1 does not decide this
score.'' The addendum:
\begin{lstlisting}[style=prompt]
---

## Multi-turn session
This edit is one step of a multi-turn editing session. Video 1 is exactly the video the editor received for this step and may already contain edits from earlier steps; it is not necessarily the original footage.

### Instructional Following for listed intents
Every listed intent is a target state the edited video should show. Some were requested for this step; others were achieved in earlier steps and should still hold. Score every one the same way, as a state shown in Video 2:
- Judge the state, not whether anything changed. If Video 2 shows the state, it counts even when Video 1 already showed it.
- Score accuracy as described in Dimension 1, and also how long the state holds: 4 it holds for the whole clip; 3 it holds for more than half of the clip, or drops out only in a few frames; 2 it holds for half of the clip or less; 1 it is not shown. Give the lower of the two.

### Undo statements (only if listed)
Each describes an earlier edit that is no longer wanted: it should hold in Video 1 and must no longer hold in Video 2. Score under Instructional Following: 4 it no longer holds and the area looks like the original scene (or like a listed intent on the same element); 3 it no longer holds apart from minor residue; 2 it partly still holds; 1 it still holds.

### Edit Exclusivity
Judge only content that no listed intent or undo statement is about: it must stay as in Video 1. Changes to what the listed intents describe are judged under Instructional Following, not here.
\end{lstlisting}

The payload lists every active intent as a target state: first the ones this attempt executes (with
any exhausted extras), then the preserved ones, without marking which ones were requested in this turn.
An exhausted intent that is not part of the attempt is left out. Then come the undo statements and the
output fields. Video 1 is the frames the editor received and Video 2 the candidate,
both sampled at 3.92\,fps (8 frames of a 49-frame clip). \agentname{} never lists undo
statements: entry selection excludes entries with an unwanted edit.

\subsection{Intent Compiler of the Single-Turn Agent}
\begin{lstlisting}[style=prompt]
You turn a video-editing request into edit intents.

The request gives the edits for one source video as a list of instructions; a single instruction may describe several edits. You do not see the video. Each intent is later planned and executed by a video editor and checked on its own by a video judge. The guidance below serves that purpose; where a case is not covered, do what serves it best. Return only JSON matching the supplied schema, keep every instruction index, and do not resolve an ambiguous target by guessing: return blocked_ambiguity instead.

What the user wants
- Do not add edits the user did not ask for.

Splitting into intents
- Aim for one intent per thing the user asked to change, so that each can be judged on its own. Changing an existing wall's color and its texture is two intents; "add a blue brick wall" or "replace the SUV with a white sedan" is one, because the appearance describes the new element. Several elements in one instruction usually mean several intents, and separate instructions should not be merged just to have fewer.
- Write each intent's "text" as a self-contained description of the result, understandable without the request, for example "A table stands to the right of the window and to the left of the lamp."
- Describe what the edited video shows, in the present tense, never the edit itself: "The fish has no dorsal fin.", not "The dorsal fin is removed."
- Carry over the new information the user gave about that edit: where it goes, how it relates to other things (for instance that the particles must not cover the flowers), its size, material or placement.
- Avoid wording that only makes sense relative to the request or the editing, such as "undo", "instead" or "again".

Fields for the planner
- "instruction_index": the instruction the intent comes from. "target": the element it changes, as the source video shows it. "operation": a short verb for the change (restyle, replace, add, remove, recolor, ...). "desired_state": the target's new appearance in a few words.

Constraints
- List in "preservation_constraints" what the instruction only asks to keep as it would stay anyway, such as "keep the lamp's light unchanged" or "leave the other cars alone". Anything that adds new information belongs in an intent instead.
\end{lstlisting}

The payload lists the written instructions with their indices and the output schema. The model returns a
status (ok or blocked\_ambiguity, with a reason) and the intents, each with its instruction index, text,
target, operation, desired state, and preservation constraints. Every instruction index must appear.

\subsection{Single-Turn Verifier (Base Prompt)}
\label{app:prompt-verifier}
The three dimensions, their criteria, and their rubrics are adapted from the reward prompt of
VEFX-Bench, the source of our clips (App.~\ref{sec:bench-clips}).
\begin{lstlisting}[style=prompt]
You are an expert evaluator assessing the quality of AI-generated video edits. You will be provided with two videos:
- **Video 1**: The Original Video (before editing)
- **Video 2**: The Edited Video (after editing)

The editing intents are listed in the request. Your task is to evaluate the Edited Video across THREE independent dimensions. Each dimension is scored on a 1-4 integer scale. **Scores across dimensions are independent** -- a failure in one dimension must NOT affect scores in another.

---

## Step 1: Survey the differences before you score anything
Write the survey first. Walk the frame region by region and list every place where Video 2 differs from Video 1: the requested edits, and everything else. Hunt for differences; do not summarise, and do not describe what stayed the same. Small, local and short-lived differences count and are the ones most often missed: leftover fragments of a removed object, invented content in blurred areas, a shift in sharpness or depth of field, colour or saturation drift on an object nobody asked you to change, flicker confined to a few frames. For each difference say where it is, when it happens, and whether an instruction asked for it.

---

## Dimension 1: Instructional Following (IF)
**Core question:** Does the edited video accurately reflect the semantic requirements of the editing instruction?

Score every intent separately. Do not let success on one intent compensate for failure on another.

Evaluation criteria:
- Object replacement: If the instruction says "replace apple with orange," did the model actually generate an orange (not a lemon or tomato)?
- Action/attribute changes: If the instruction involves motion or attribute changes (e.g., "make it night"), was this correctly executed?
- Completeness: Were ALL parts of the instruction addressed, not just partial execution?

Scoring rubric:
- **4 (Perfect):** The edit precisely and completely executes all instructions. Object categories, attributes (color, shape), actions, and styles all match the instruction with no ambiguity.
- **3 (High):** The main instruction was executed, but minor details deviate. E.g., instruction asks for "red sports car" but a "red truck" was generated -- the main concept "red car" is correct.
- **2 (Low):** The main instruction was partially executed but with significant deviations, or completely irrelevant modifications were made.
- **1 (Failed):** The edit has no relation to the instruction. E.g., instruction asks for "night scene" but the video remains daytime, or no change occurred at all.

**Important notes:**
- If the edit instruction asks for a camera perspective change (e.g., "shift to high angle") and the video shows no actual perspective change, score 1.
- If the instruction asks for adding/increasing objects and no new objects appear, score 1.
- A video that looks identical to the original (no edit happened) always scores 1 for IF.

---

## Dimension 2: Render Quality (RQ)
**Core question:** What is the visual and temporal quality of the edited video?

Evaluation criteria:
- Naturalness and clarity: Are all parts of the video natural and sharp? Any blurriness, noise, or artifacts?
- Physical plausibility: Does object motion obey physics? Any flickering, jittering, objects disappearing/morphing unexpectedly?
- Temporal consistency: Is the video smooth frame-to-frame? Any sudden jumps, abrupt texture/color changes between frames?

Scoring rubric:
- **4 (Excellent):** Video clarity is very high with no visible defects, or only extremely minor artifacts detectable on very close inspection. Object motion fully obeys physics, smooth and natural. Visual quality is on par with or better than the original.
- **3 (Medium):** Some quality degradation exists (e.g., slight blurring, localized flickering), but all objects remain clearly identifiable. The video's overall structure is intact despite imperfections.
- **2 (Poor):** Significant quality degradation with obvious artifacts, distortion, or frame-to-frame inconsistency. Some object outlines deform, motion appears unnatural, affecting viewing experience.
- **1 (Unusable):** Video quality completely breaks down. Objects are severely deformed or unrecognizable, serious physics violations (e.g., person walking through walls, objects shattering spontaneously), heavy noise or complete blur.

**Important notes:**
- A sudden scene transition mid-video (e.g., white background abruptly becoming a construction site) counts as a physics/consistency violation -- score <= 3.
- If the edit did NOT happen (original is preserved), RQ can still be high if the video itself looks fine -- evaluate the video's visual quality independently.
- Evaluate temporal artifacts carefully: a single frame of flickering is minor (score 3), persistent warping or morphing is severe (score 1-2).

---

## Dimension 3: Edit Exclusivity (EE)
**Core question:** Did the model ONLY perform the specified edit, without making unintended changes to other parts of the video?

Evaluation criteria:
- Over-editing: When editing a foreground object, did the background, lighting, or other unrelated objects change?
- Scene consistency: Are pixels, textures, and structures in non-edited regions preserved?
- Camera trajectory: Was the original camera movement preserved? (Changing camera motion when not instructed is over-editing.)
- Identity preservation: Do unedited people maintain their facial features, expressions, and body movements?

Scoring rubric:
- **4 (Perfect):** Strict exclusivity maintained. Only the target region specified by the instruction changed. All other regions (background, unrelated objects) remain identical to the original. Tiny pixel-level differences invisible to the eye are acceptable.
- **3 (Medium):** Visible over-editing occurred. Non-target areas show noticeable changes, but overall scene layout and unrelated object consistency are still preserved. E.g., replaced a cup on a table but the table style also changed, or a background window disappeared.
- **2 (Poor):** The overall scene or multiple unrelated objects changed significantly.
- **1 (Complete failure):** No exclusivity at all. The entire video looks like a completely new video. The surrounding scene changed drastically, or more than three unrelated objects underwent serious alterations.

**Important notes:**
- Camera trajectory changes (when not instructed) are over-editing -- if the original video had camera motion and the edited video is static (or vice versa), penalize EE.
- If the edit failed (IF=1) but the rest of the video also changed, EE should still be scored low.

---

## Explanations
4 is the full score. Every score needs an evidence-based explanation. A score of 4 means no defect; say briefly why. Any score below 4 must name every concrete defect that cost points: what is wrong, where in the frame, and when it happens. For Edit Exclusivity, also say where each unintended change sits relative to the requested edits, so it can be traced back.
\end{lstlisting}